\documentclass{article}
\usepackage{iclr2027_conference,times}

\usepackage{amsmath,amsfonts,bm}

\def\eqref#1{equation~\ref{#1}}

\def\1{\bm{1}}

\DeclareMathAlphabet{\mathsfit}{\encodingdefault}{\sfdefault}{m}{sl}
\SetMathAlphabet{\mathsfit}{bold}{\encodingdefault}{\sfdefault}{bx}{n}

\newcommand{\E}{\mathbb{E}}

\newcommand{\normltwo}{L^2}

\newcommand{\method}{LILIE}
\newcommand{\xlsize}{XL}
\newcommand{\gsize}{g}

\usepackage{titletoc}
\usepackage{hyperref}
\usepackage{cleveref}
\usepackage{url}
\usepackage{amsmath}
\usepackage{amssymb}

\title{A Language-Informed Loss for \\ Visual Generation}

\author{%
Suchir Agarwal \quad
Keshigeyan Chandrasegaran \quad
Ziyu Chen \quad
Kyle Sargent \\
\bfseries Juan Carlos Niebles \quad
Li Fei-Fei \quad
Jiajun Wu \\[0.5em]
Stanford University
}

\usepackage{adjustbox}
\usepackage{booktabs}
\usepackage{multirow}
\usepackage{algorithm}
\usepackage{algpseudocode}
\usepackage{wrapfig}
\usepackage{needspace}

\usepackage{xcolor}
\definecolor{okblack}{HTML}{000000}
\definecolor{okorange}{HTML}{E69F00}
\definecolor{oksky}{HTML}{56B4E9}
\definecolor{okgreen}{HTML}{009E73}
\definecolor{okyellow}{HTML}{F0E442}
\definecolor{okblue}{HTML}{0072B2}
\definecolor{okvermillion}{HTML}{D55E00}
\definecolor{okpurple}{HTML}{CC79A7}

\usepackage{array}
\usepackage{colortbl}
\usepackage{pifont}
\newcommand{\cmark}{\textcolor{okgreen}{\ding{51}}}
\newcommand{\xmark}{\textcolor{okvermillion}{\ding{55}}}
\newlength\savewidth\newcommand\shline{\noalign{\global\savewidth\arrayrulewidth
  \global\arrayrulewidth 1pt}\hline\noalign{\global\arrayrulewidth\savewidth}}
\newcommand\midline{\noalign{\global\savewidth\arrayrulewidth
  \global\arrayrulewidth 0.5pt}\hline\noalign{\global\arrayrulewidth\savewidth}}
\newcommand{\tablestyle}[2]{\setlength{\tabcolsep}{#1}\renewcommand{\arraystretch}{#2}\centering\footnotesize}
\newcolumntype{x}[1]{>{\centering\arraybackslash}p{#1pt}}
\usepackage{etoolbox}
\makeatletter
\pretocmd{\WF@putfigmaybe}{%
  \ifinner\else
    \Needspace{\WF@size}%
  \fi
}{}{\PackageError{paper}{Could not install wrap-figure page guard}{Check the wrapfig package version.}}
\makeatother
\AtBeginEnvironment{table}{\setlength{\belowcaptionskip}{\baselineskip}}
\usepackage{listings}
\usepackage{graphicx}

\colorlet{codeblue}{okgreen}
\colorlet{codesign}{okblue}
\colorlet{codefunc}{okpurple}
\lstdefinelanguage{PythonFuncColor}{
  language=Python,
  keywordstyle=\color{okblue}\bfseries,
  commentstyle=\color{codeblue},
  stringstyle=\color{okorange},
  showstringspaces=false,
  deletekeywords={lambda},
  basicstyle=\ttfamily\small,
  literate=
    {*}{{\color{codesign}* }}{1}
    {-}{{\color{codesign}- }}{1}
    {+}{{\color{codesign}+ }}{1}
    {/}{{\color{codesign}/ }}{1}
    {preprocess}{{\color{codefunc}preprocess}}{1}
    {encode(}{{\color{codefunc}encode}(}{7}
    {encode_for_loss}{{\color{codefunc}encode\_for\_loss}}{1}
    {sample_t}{{\color{codefunc}sample\_t}}{1}
    {l2_loss}{{\color{codefunc}l2\_loss}}{1}
    {net_fn}{{\color{codefunc}net}}{1}
    {rms}{{\color{codefunc}rms}}{1}
    {randn_like}{{\color{codefunc}randn\_like}}{1}
    {cross_entropy}{{\color{codefunc}cross\_entropy}}{1}
    {vlm}{{\color{codefunc}vlm}}{1}
}
\iclrpreprintcopy
\begin{document}

\maketitle
\begingroup
\renewcommand{\thefootnote}{}
\footnotetext{Correspondence to \texttt{suchir@cs.stanford.edu}.}
\endgroup

\begin{abstract}
End-to-end visual generation with pixel-space loss has recently achieved competitive performance without the need for learned image tokenizers. In this work, we study whether language-informed supervision can provide a richer training signal than pixel loss for visual generation. We introduce \textbf{\method{}} (pronounced ``lily''), a \textbf{L}anguage-\textbf{I}nformed and \textbf{L}ossless \textbf{I}mage \textbf{E}ncoder. \method{} is an invertible image transform trained such that a vision-language model can recover captions from noised features. The encoder uses a normalizing flow transformer trained jointly with a vision-language model via next token prediction. We then use the encoder to extract a simple language-informed loss for text-to-image generation. We train JiT models with \method{} loss, reducing gFD by 14 points {(66.84 $\to$ 53.01)} against a pixel-only baseline on the GPIC dataset. Our code, encoders, and generative models are available at \href{https://suchiragarwal.com/lilie}{\textcolor{okblue}{suchiragarwal.com/lilie}}.

\end{abstract}

\begin{figure}[!htb]
  \centering
  \includegraphics[width=0.91\linewidth]{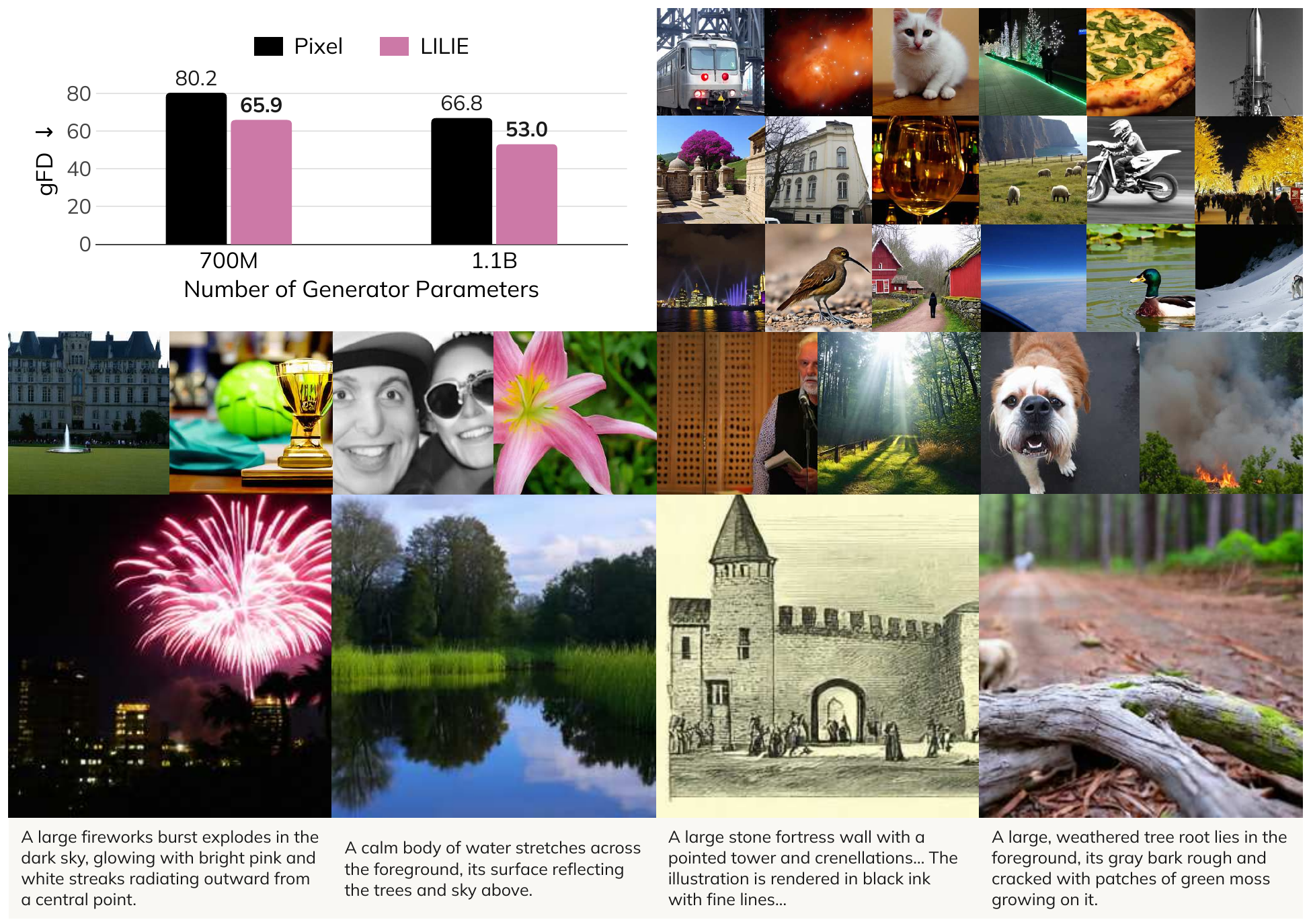}
  \vspace{-5pt}
  \caption{\textbf{Generation quality with \method{}.} Upper left: Pixel and \method{} supervision at two generator sizes evaluated on GPIC Test-50K with 50 Euler steps. CFG scale is selected using GPIC Validation-50K. All models are trained for one epoch on GPIC Full~\citep{chandrasegaran2026gpic}. Mosaic: selected $256\!\times\!256$ samples from the 1.1B diffusion model with \method{} supervision. Four samples include prompt excerpts. We include randomly selected samples from GPIC Test-50K in Appendix~\ref{sec:supp:samples}.}
  \label{fig:main:fig1}
\end{figure}

\clearpage

\section{Introduction}

Image generators are typically built in two stages. A tokenizer first compresses images into a learned latent space, and a generative model is then trained in that space. Latent diffusion \citep{rombach2021highresolution,Peebles2022DiT}, autoregressive \citep{sun2024llamagen}, and mixed-modal \citep{zhou2024transfusion} generators all follow this recipe.
By reducing the dimensionality of the modeling problem, tokenization makes generation more tractable and lowers computational cost. However, the lossy tokenizer bounds achievable quality, particularly for fine-grained content (e.g.\ rendered text and human faces) \citep{yao2025vavae,yang2025detok}. Recent work removes the first stage entirely. JiT \citep{li2025jit} trains a plain transformer end-to-end on raw pixels and achieves competitive performance with two-stage systems without training a tokenizer at all. Pixels are hence an appealing output space, but we ask whether generation quality can be improved by extracting a richer training signal.

\begin{wrapfigure}[17]{l}{0.35\textwidth}
\vspace{-\intextsep}
\centering
\includegraphics[width=\linewidth]{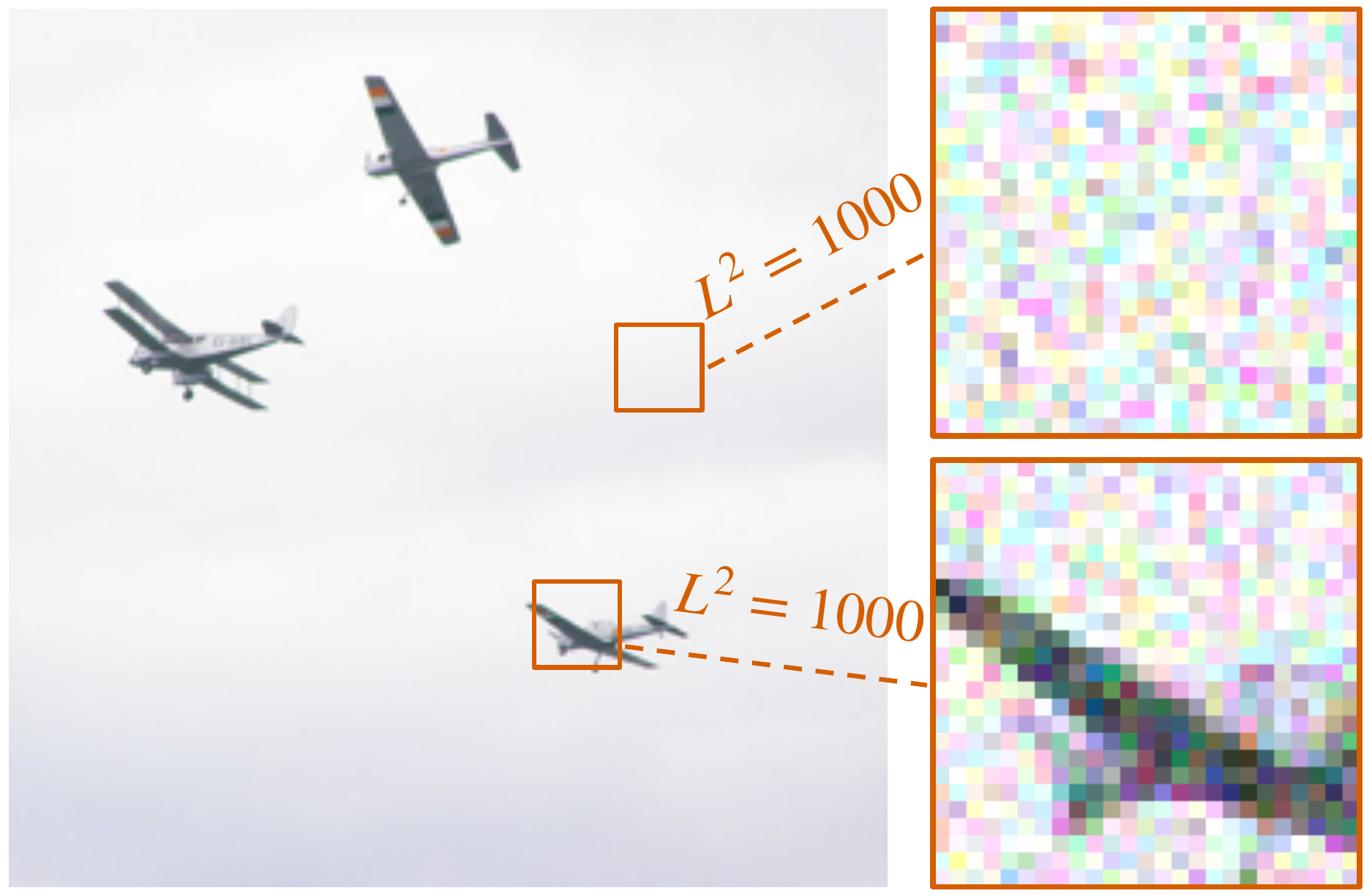}
\vspace{-10pt}
\caption{Two perturbed image patches (additive Gaussian noise) with equal pixel Euclidean distance of 1000 from the original patches. Caption: ``Three airplanes fly in the sky against a cloudy backdrop.''}
\label{fig:main:example-errors}
\end{wrapfigure}

Pixel loss measures squared Euclidean distance between predicted and target pixels. However, errors of equal pixel distance can vary in underlying semantic content. Figure~\ref{fig:main:example-errors} shows two perturbations, each with pixel distance of 1000 from the original patches. One perturbation degrades a subject while the other alters a background cloud, but pixel loss treats them the same. We hypothesize that language may help address this gap. Prior work suggests that language provides an inherently semantic signal \citep{Hockett1960TheOO,Mikolov2013DistributedRO,BARBIERI2020104100}. Motivated by the semantic gap between language and vision \citep{smeulders2000content,collell-moens-2016-image}, we study whether distances in a visual space shaped by language can provide a richer training signal to supervise end-to-end visual generation than pixel loss alone.

\begin{wrapfigure}[19]{r}{0.48\textwidth}
\vspace{-\intextsep}
\setlength{\abovecaptionskip}{4.32pt}
\centering
\includegraphics[width=\linewidth]{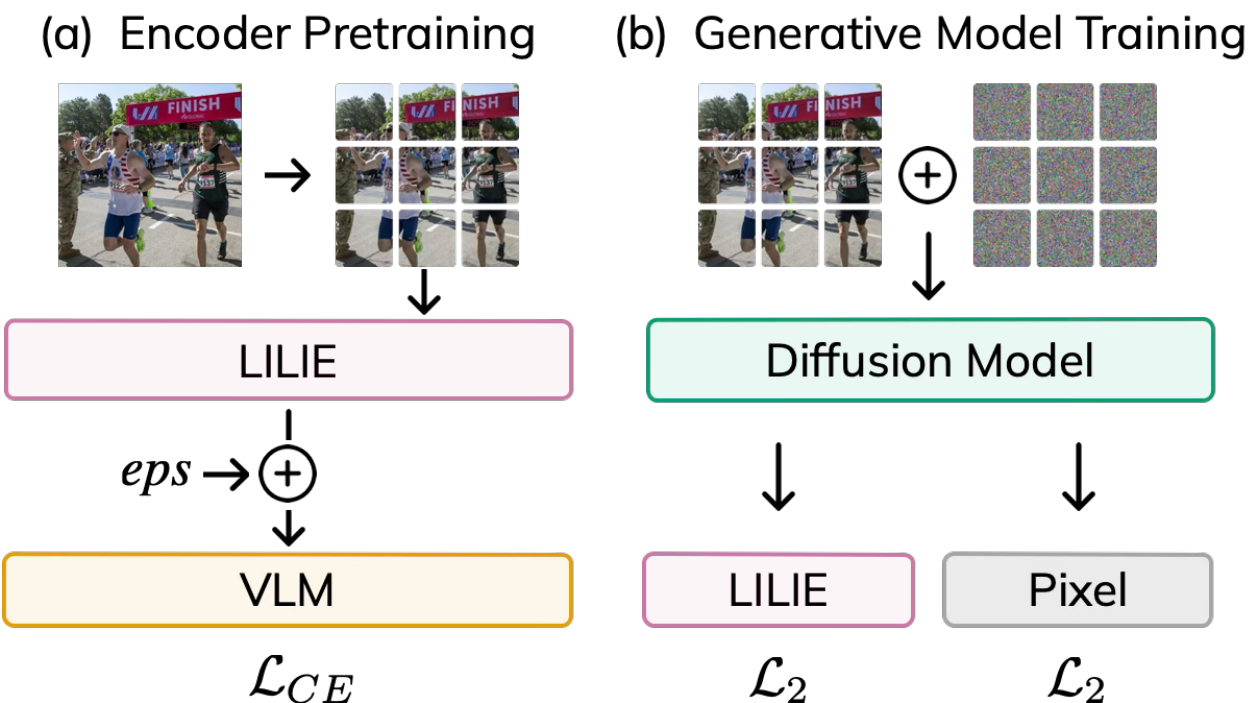}
\vspace{-6pt}
\caption{\textbf{Training and using \method{}.}
(a) We jointly train \method{} and VLM LoRA parameters on noise-corrupted features with next token prediction.
(b) The diffusion model trains on Pixel and \method{} $\normltwo$ losses, comparing its
prediction with the target image in pixel space and through the learned encoder.\strut}
\label{fig:main:llsvr-overview}
\end{wrapfigure}

We introduce \method{}, a Language-Informed and Lossless Image Encoder. \method{} is an invertible transform from $\mathbb R^{H \times W \times 3} \to \mathbb R^{H \times W \times 3}$, trained such that a vision-language model can caption the output after it is corrupted by Gaussian noise. We build \method{} on a transformer normalizing flow architecture \citep{zhai2024tarflow,lu2025bidirectional}, keeping the invertible architecture but replacing the generation objective with a captioning objective. Our method is trained with next token prediction over ground-truth image captions. The encoder learns to arrange its output so that caption-relevant visual information survives the corruption, while invertibility guarantees that no visual information is lost. After training, we discard the VLM and keep \method{} as a simple standalone image transform.

We finally train a family of JiT-T2I \citep{li2025jit,ma2026pixelgen} models on GPIC \citep{chandrasegaran2026gpic}, a large scale open dataset for text-to-image generation. We first pretrain pixel-space and perceptual loss baselines to provide references on the GPIC corpus. We then study the impact of our training signal with \method{} on generation quality by performing comprehensive ablations. We freeze the encoder, apply it to the predicted and target images, and add squared error in \method{}-space to the standard pixel-space flow-matching loss \citep{lipman2023flow}. For our 1.1B parameter JiT-T2I-\gsize{}, the pixel-only baseline reaches a gFD-DINOv2 \citep{stein2023exposing} of 66.84 on the GPIC test set. Replacing the pixel loss entirely with \method{} loss improves this to 55.78, so language supervision alone is already enough to learn a feature space that improves pixel generation. Combining \method{} and pixel losses further improves gFD to 53.01.

To summarize, we propose a language-informed loss that improves pixel-space visual generation quality, and we hope our work encourages the community to adopt language-informed losses for training visual generative models. We highlight our main contributions below. 
\begin{itemize}
    \item We pretrain language-informed and lossless image encoders.
    \item We pretrain generative models with perceptual losses (LPIPS, DINOv2, SigLIP 2) on the GPIC corpus to provide baseline references.
    \item We pretrain generative models with supervision in \method{}-space and achieve a 14-point reduction in gFD over a pixel-only baseline.
\end{itemize}

\section{Related Work}
\label{sec:main:related-work}

\paragraph{Pixel-space diffusion.}
Pixel-space diffusion generates images without a learned image tokenizer. Cascaded diffusion models, including Imagen, generate images at progressively higher resolutions \citep{ho2022cascaded,saharia2022imagen}, while Simple diffusion and Recurrent Interface Networks support high-resolution generation without cascades \citep{hoogeboom2023simple,jabri2023rin}. More recent approaches include cascaded flow matching in PixelFlow \citep{chen2025pixelflow}, neural field prediction within image patches in PixNerd \citep{wang2025pixnerd}, and joint patch-level and pixel-level processing in PixelDiT \citep{yu2026pixeldit}. JiT predicts clean images directly from noisy pixels \citep{li2025jit}. Pixel MeanFlow extends clean-image prediction to one-step generation with a MeanFlow objective \citep{lu2026pixelmeanflow,geng2025meanflow}. We build on JiT, studying the training loss while keeping the generator architecture fixed.

\paragraph{Feature-space losses for generation.}
Perceptual losses compare images through neural features, providing supervision beyond pixel-wise error \citep{Johnson2016Perceptual,zhang2018perceptual}. PixelGen applies this approach by augmenting a pixel-space flow-matching loss with LPIPS and DINOv2 losses \citep{ma2026pixelgen}. REPA instead aligns the denoiser's hidden states with features from a pretrained visual encoder \citep{yu2025repa}. PixelREPA finds that directly applying this alignment to JiT can reduce sample quality and diversity, and introduces a masked adapter that improves generation \citep{shin2026pixelrepa}. VA-VAE and representation autoencoders also use pretrained visual features to construct latent spaces for diffusion \citep{yao2025vavae,zheng2025rae}. In this work, we learn a language-supervised and lossless representation for the loss while keeping generation in pixel space.

\paragraph{Invertible networks.}
Prior normalizing flow models such as Real NVP and Glow learn invertible transformations that enable exact likelihood evaluation and sampling \citep{dinh2017realnvp,kingma2018glow}. TARFlow extends this approach with autoregressive transformers operating on image patches \citep{zhai2024tarflow}. Invertible architectures have also been used for representation learning as in i-RevNet \citep{jacobsen2018irevnet}. IB-INN uses small input perturbations to form an information-bottleneck objective for invertible generative classifiers \citep{ardizzone2020ibinn}. \method{} uses the improved TARFlow architecture introduced in BiFlow \citep{lu2025bidirectional}. We train this bijection through caption prediction from corrupted features and use its learned feature space to measure errors when training a T2I generator.

\paragraph{Language-supervised visual representations.}
CLIP learns transferable visual features through contrastive image-text supervision \citep{radford2021clip}, while SigLIP 2 combines image-text alignment with captioning and self-supervised objectives \citep{tschannen2025siglip}. VirTex and CapPa demonstrate that caption prediction itself can learn effective visual representations \citep{desai2021virtex,tschannen2023cappa}. Language has also been used to inform image tokenization. QLIP and UniTok combine reconstruction and image-text alignment objectives to learn discrete representations \citep{zhao2025qlip,ma2025unitok}. TexTok conditions image encoding and decoding on captions, improving reconstruction and downstream generation quality \citep{zha2025language}. \method{} uses captions as prediction targets for corrupted image features. Caption cross-entropy is our only training loss.

\newpage
\section{Methodology}
\label{sec:main:method}

Pixel-space losses treat errors across pixels equally, even when they differ in relevance to image content. Feature-space losses instead compare images in a feature space that emphasizes certain visual properties. Here, we propose a method for learning such a feature space that is lossless and supervised by language.

\method{} learns a bijection $\mathcal E_\theta : \mathbb R^{H \times W \times 3} \to \mathbb R^{H \times W \times 3}$ from images to equally-shaped features. We train the encoder such that a finetuned VLM can recover an image's caption from features corrupted with Gaussian noise. This corruption procedure encourages semantic information to remain recoverable under corruption, and invertibility prevents representations from discarding any visual information.

Our complete method has two stages. First, we learn an image encoder $\mathcal E_\theta$ from image-caption pairs. We then freeze the encoder and use $\normltwo$ in \method{}-space to augment pixel-space loss for training a text-to-image generative model.

\subsection{\method{}: A Language-Informed and Lossless Image Encoder}

\paragraph{Learning an invertible map.}
We parametrize $\mathcal E_\theta$ using the improved TARFlow architecture from BiFlow \citep{lu2025bidirectional, zhai2024tarflow}. Unlike prior applications, we do not use this architecture to train a normalizing flow density model. We use it for its invertibility and change the training objective.

\begin{figure}[!htb]
\centering
\includegraphics[width=\linewidth]{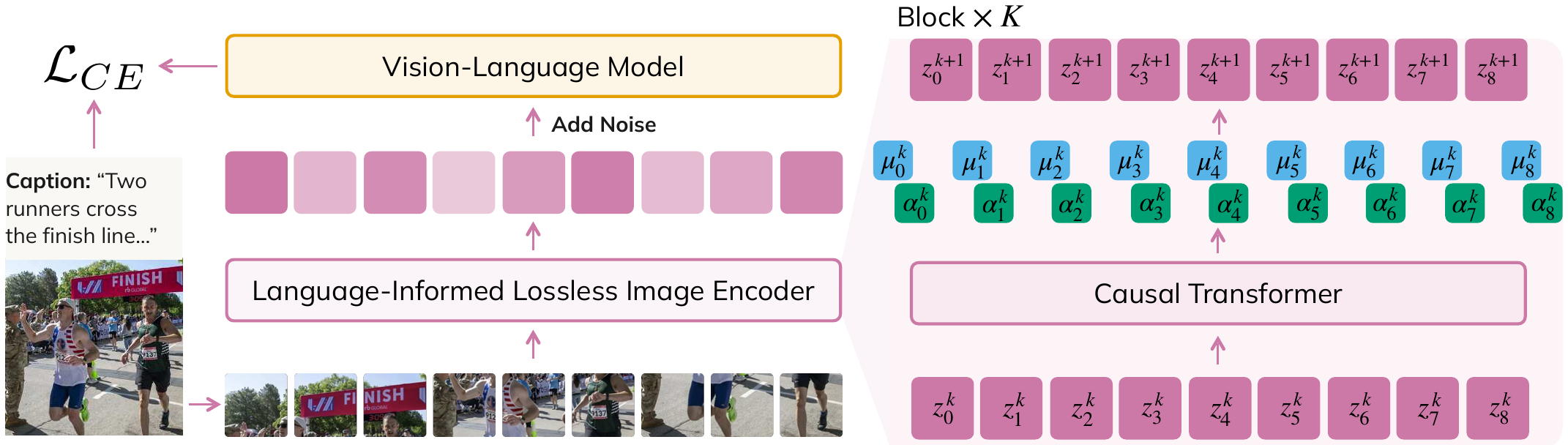}
\vspace{-10pt}
\caption{
\textbf{\method{} training diagram.} We expand on Figure~\ref{fig:main:llsvr-overview}(a) to include TARFlow internals. We provide additional details on affine coupling and transformer internals in Appendix~\ref{sec:supp:encoder-architecture}.
}
\label{fig:main:llsvr-training}
\end{figure}

Figure~\ref{fig:main:llsvr-training} and Algorithm~\ref{alg:llsvr-training} describe the training procedure to learn $\mathcal E_\theta$. Each image is mapped through $\mathcal E_\theta$, rescaled to a per-image RMS of $1/2$, corrupted with additive white Gaussian noise, and captioned by a LoRA-finetuned VLM \citep{hu2022lora}. The training signal is cross-entropy of the VLM predictions against the ground-truth GPIC caption. We backpropagate to jointly update $\mathcal E_\theta$ and the VLM's LoRA parameters. The RMS target of $1/2$ approximately matches the RMS of natural images in the GPIC corpus, keeps \method{} feature scales in-domain for the VLM, and prevents $\mathcal E_\theta$ from enlarging all features to compensate for the corruption. We also track an EMA estimate of the raw features' second moment, averaged over all features, which we use for downstream rescaling. After training, we discard the VLM and retain only the encoder and the second moment estimate.

\begin{figure}[!htb]
\centering
\begin{minipage}[t]{0.49\linewidth}
\vspace{0pt}
\begin{algorithm}[H]
\caption{\method{} training.}
\label{alg:llsvr-training}
\small
\textbf{Input:} Image-caption batches \texttt{data}, initial invertible
\texttt{encode}, pretrained \texttt{vlm} with LoRA, noise level
\texttt{sigma}, EMA decay \texttt{beta}.\par
\textbf{Output:} Trained \texttt{encode} and second moment \texttt{m2}.
\begin{lstlisting}[basicstyle=\fontsize{8}{9.6}\ttfamily\selectfont,escapeinside={(*@}{@*)}]
for image, caption in data:
    x = preprocess(image) # (*@\hyperref[sec:supp:algorithm-details]{Appendix~\ref*{sec:supp:algorithm-details}}@*)
    z = encode(x)
    u = z / (2 * rms(z))  # per image
    y = u + sigma * randn_like(u)
    loss = cross_entropy(vlm(y), caption)
    update_params(loss, encode, vlm.lora)

    # Track the second moment (EMA)
    q = mean(square(z.detach()))
    m2 = ema_update(m2, q, beta)

return encode, m2
\end{lstlisting}
\end{algorithm}
\end{minipage}\hfill
\begin{minipage}[t]{0.49\linewidth}
\vspace{0pt}
\begin{algorithm}[H]
\caption{JiT training loss.}
\label{alg:t2i-training}
\small
\textbf{Input:} Frozen \texttt{encode} and \texttt{m2} from
Algorithm~\ref{alg:llsvr-training}; target \texttt{x}, prediction
\texttt{x\_pred}, time weight \texttt{w(t)}, loss weight \texttt{lambda}.\par
\textbf{Output:} Generator training loss.
\begin{lstlisting}[basicstyle=\fontsize{8}{9.6}\ttfamily\selectfont]
# Fixed rescale, preserves invertibility
def encode_for_loss(x):
    return encode(x) / (2 * sqrt(m2))

z = encode_for_loss(x)
z_pred = encode_for_loss(x_pred)

loss_pixel = l2_loss(x - x_pred)
loss_lilie = l2_loss(z - z_pred)
loss = w(t) * (loss_pixel
               + lambda * loss_lilie)

return loss
\end{lstlisting}
\end{algorithm}
\end{minipage}
\end{figure}

\subsection{Training a T2I Generative Model with \method{}}

We now freeze the \method{} encoder $\mathcal E_\theta$ and use it to supervise pixel-space text-to-image generation.

\paragraph{Base objective.}
We build on JiT-T2I \citep{li2025jit,ma2026pixelgen,chandrasegaran2026gpic}, a plain transformer architecture that denoises pixels directly, conditioned on caption embeddings from a frozen Qwen3-1.7B text encoder \citep{qwen3}. 
The JiT $x$-prediction loss is
\[
\mathcal L_\text{Pixel} = \E [ w(t) \Vert x - \hat x \Vert_2^2], \quad w(t) = \frac{1}{\max (1-t, t_\text{clip})^2},
\]
the velocity-matching objective rewritten in $x$-prediction form \citep{li2025jit}. The caption is dropped for a random 10\% of samples to learn the unconditional path and support classifier-free guidance at sampling time \citep{ho2022classifier}. 

\paragraph{Augmenting with \method{}.}
We freeze $\mathcal E_\theta$ and additionally score the prediction in \method{}-space. Analogous to the rescaling performed in encoder training, we define
\begin{equation}
\label{eq:encode-for-loss}
\tilde{\mathcal E}(x) \triangleq \frac{\mathcal E_\theta(x)}{2 \sqrt{m_2}},
\end{equation}
where $m_2$ is an EMA estimate of the second moment computed during training (Algorithm~\ref{alg:llsvr-training}). Rescaling for $\tilde{\mathcal E}$ only depends on corpus statistics, so $\tilde{\mathcal E}$ remains a bijection. We finally use $\tilde{\mathcal E}$ to construct our training objectives. We consider two objectives. The first is JiT supervision \textit{solely} in \method{}-space, and we additionally combine pixel and \method{}-space losses.
\[
\mathcal L_\text{\method{}} = \E [w(t) \Vert \tilde{\mathcal E}(x) - \tilde{\mathcal E}(\hat x) \Vert_2^2], \qquad
\mathcal L_\text{Pixel+\method{}} = \mathcal L_\text{Pixel} + \lambda \mathcal L_\text{\method{}},
\]
where $\lambda$ is a loss weighting hyperparameter. Algorithm~\ref{alg:t2i-training} summarizes the Pixel+\method{} JiT training loss.

\section{Experiments}
\label{sec:main:experiments}

\begin{table}[!t]
\caption{\textbf{Generation results on GPIC.} We compare generation quality achieved by JiT-T2I-XL (700M) and JiT-T2I-g (1.1B) models trained with different loss functions. Models train for 390,600 steps, and we report results at 781,200 steps for selected configurations. We compare Pixel, \method{}-only, and Pixel+\method{} losses at the 700M generator size. At 1.1B, we compare \mbox{(i) Pixel} loss; \mbox{(ii) standalone} feature-space losses with DINOv2, SigLIP 2, LPIPS (VGG16), and \method{} backbones; (iii) feature-space losses complementary to Pixel loss. Lower is better for gFD, higher is better for precision, recall, coverage, and values near 1 are best for density. We bold rows achieving lowest gFD within each generator size and training budget. ``GPIC compliant'' indicates compliance with the GPIC evaluation protocol. We note that the DINOv2 and SigLIP 2 models used for feature-space losses are both distilled from larger models, while VGG16 and \method{} are not distilled. DINOv2, SigLIP 2, and LPIPS losses result in degenerate generation quality when used without a Pixel loss term. \textbf{Key result:} \method{} loss yields competitive quality for end-to-end visual generation, both when used as a standalone loss and as a complement to Pixel loss.
Table~\ref{tab:supp:headline-results-full} includes selected CFG values and gFD decomposition.}

\label{tab:main:headline-results}
\centering
\small
\setlength{\tabcolsep}{4pt}
\begin{adjustbox}{max width=\linewidth}
\begin{tabular}{lclcccccc}
\toprule
\multicolumn{2}{l}{Generator} & Loss & gFD $\downarrow$ & Precision $\uparrow$ & Recall $\uparrow$ & Density $\to 1$ & Coverage $\uparrow$ & MMD $\downarrow$ \\
\midrule
\multicolumn{2}{l}{} & Pixel & 80.24 & 0.932 & 0.793 & 0.979 & 0.913 & 0.052 \\
\multicolumn{2}{l}{} & \cellcolor{okpurple!16}\method{} & \cellcolor{okpurple!16}68.88 & \cellcolor{okpurple!16}0.941 & \cellcolor{okpurple!16}0.829 & \cellcolor{okpurple!16}0.996 & \cellcolor{okpurple!16}0.930 & \cellcolor{okpurple!16}0.046 \\
\multicolumn{2}{l}{\multirow{-3}{*}{\shortstack{JiT-T2I-\xlsize{}\\700M}}} & \cellcolor{okpurple!16}Pixel+\method{} & \cellcolor{okpurple!16}\textbf{65.86} & \cellcolor{okpurple!16}\textbf{0.943} & \cellcolor{okpurple!16}\textbf{0.832} & \cellcolor{okpurple!16}\textbf{1.005} & \cellcolor{okpurple!16}\textbf{0.932} & \cellcolor{okpurple!16}\textbf{0.042} \\
\midrule
 & \cellcolor{black!7}GPIC & Pixel & 66.84 & 0.940 & 0.825 & 1.000 & 0.926 & 0.038 \\
\cline{3-9}
 & \cellcolor{black!7}compliant & & & & & & & \\
 & \xmark & DINOv2 & 3124.01 & 0.987 & 0.000 & 1.082 & 0.048 & 15.225 \\
 & \xmark & SigLIP 2 & 3161.15 & 0.952 & 0.000 & 1.172 & 0.050 & 14.817 \\
 & \cmark & LPIPS (VGG16) & 158.45 & 0.922 & 0.651 & 1.053 & 0.848 & 0.174 \\
 & \cellcolor{okpurple!16}\cmark & \cellcolor{okpurple!16}\method{} & \cellcolor{okpurple!16}55.78 & \cellcolor{okpurple!16}0.947 & \cellcolor{okpurple!16}0.855 & \cellcolor{okpurple!16}1.028 & \cellcolor{okpurple!16}0.942 & \cellcolor{okpurple!16}0.031 \\
\cmidrule(lr){2-9}
 & \xmark & Pixel+DINOv2 & \textbf{47.88} & \textbf{0.951} & \textbf{0.864} & \textbf{1.051} & \textbf{0.950} & \textbf{0.023} \\
 & \xmark & Pixel+SigLIP 2 & 52.55 & 0.956 & 0.847 & 1.064 & 0.949 & 0.026 \\
 & \cmark & Pixel+LPIPS & 48.65 & 0.950 & 0.858 & 1.035 & 0.943 & 0.022 \\
\multirow{-10}{*}{\shortstack{JiT-T2I-\gsize{}\\1.1B}} & \cellcolor{okpurple!16}\cmark & \cellcolor{okpurple!16}Pixel+\method{} & \cellcolor{okpurple!16}53.01 & \cellcolor{okpurple!16}0.947 & \cellcolor{okpurple!16}0.857 & \cellcolor{okpurple!16}1.029 & \cellcolor{okpurple!16}0.943 & \cellcolor{okpurple!16}0.028 \\
\midrule
\multicolumn{9}{l}{\textit{390,600 steps above. 781,200 steps below.}} \\
\midrule
\multirow{5}{*}{\shortstack{JiT-T2I-\gsize{}\\1.1B}} & \cmark & Pixel & 49.20 & 0.951 & 0.871 & 1.025 & 0.943 & 0.025 \\
 & \xmark & Pixel+DINOv2 & \textbf{38.29} & \textbf{0.959} & \textbf{0.883} & \textbf{1.069} & \textbf{0.957} & \textbf{0.018} \\
 & \xmark & Pixel+SigLIP 2 & 40.27 & 0.956 & 0.881 & 1.066 & 0.959 & 0.019 \\
 & \cmark & Pixel+LPIPS & 39.35 & 0.952 & 0.882 & 1.043 & 0.953 & 0.017 \\
 & \cellcolor{okpurple!16}\cmark & \cellcolor{okpurple!16}Pixel+\method{} & \cellcolor{okpurple!16}40.18 & \cellcolor{okpurple!16}0.954 & \cellcolor{okpurple!16}0.886 & \cellcolor{okpurple!16}1.058 & \cellcolor{okpurple!16}0.959 & \cellcolor{okpurple!16}0.020 \\
\bottomrule
\end{tabular}

\end{adjustbox}
\end{table}

We now study the impact of \method{} on pixel-space text-to-image generation quality.

\subsection{Experimental Setup}

\paragraph{Data.} 
All models are trained on the GPIC corpus of image-caption pairs \citep{chandrasegaran2026gpic} at $256 \times 256$ resolution. Following the experimental setup from GPIC, we train both the encoder and diffusion models over one epoch on GPIC Full (100M train examples) at a global batch size of 256, yielding 390,600 steps. Every image in the GPIC corpus is paired with a caption from one of four length classes (tag, short, medium, long), which we match with a corresponding prompt for \method{} training (Appendix~\ref{sec:supp:prompts}).

\paragraph{Encoders.} 
Unless otherwise stated, the encoder is \method{}-S, a 32M parameter improved TARFlow \citep{lu2025bidirectional} with $K=4$ blocks, $L=4$ layers per block, bottleneck dimension $H=384$, and patch size 16. We use SmolVLM-256M \citep{marafioti2025smolvlm}, which we finetune with rank 64 LoRA. We choose SmolVLM-256M as it is a compact openly available VLM, at $8\times$ the size of our encoder. Training \method{}-S takes approximately 40 hours on TPU v5e-32. We detail the full configuration in Table~\ref{tab:supp:config-llsvr}.

\paragraph{Generative models.}
We train JiT-T2I-\xlsize{} and -\gsize{} models \citep{li2025jit, ma2026pixelgen} with 700M and 1.1B parameters, respectively. Both condition on a frozen Qwen3-1.7B text encoder \citep{qwen3}. We primarily compare three objectives: Pixel, \method{}, and Pixel+\method{}, while holding the generator configuration constant. Our main results train for one epoch on GPIC Full (390,600 steps), and ablations use 200K steps. Training JiT-T2I-\xlsize{} and -\gsize{} takes approximately 27 and 40 hours on TPU v6e-16, respectively. We detail the full configurations in Table~\ref{tab:supp:config-jit}.

\paragraph{Perceptual loss baselines.} Recent work has shown that perceptual supervision with LPIPS and DINOv2 significantly benefits pixel-space diffusion generation quality \citep{ma2026pixelgen}. These methods follow our general paradigm of combining pixel and feature-space losses, so we pretrain JiT models with perceptual supervision to provide reference baselines. We consider LPIPS with VGG16 \citep{zhang2018perceptual} and perceptual losses in DINOv2 ViT-B/14 \citep{oquab2023dinov2} and SigLIP 2 ViT-B/16@256px \citep{tschannen2025siglip} feature spaces. We pretrain each baseline using a combination of pixel-space loss with a single feature-space loss, and we follow the setup from PixelGen \citep{ma2026pixelgen} while disabling REPA feature alignment. SigLIP 2 provides a language-aligned baseline, and we configure it in an identical manner to P-DINO in PixelGen using cosine dissimilarity. We additionally evaluate each feature space alone, without any pixel-space loss, using $\normltwo$ distance for DINOv2 and SigLIP 2. We provide full implementation details in Appendix~\ref{sec:supp:baselines}.

\paragraph{Evaluation. } We report Fr\'echet distance (gFD), precision, recall, density, and coverage (PRDC) \citep{Kynkaanniemi2019,ferjad2020icml}, and squared maximum mean discrepancy (MMD) all in the DINOv2 representation space \citep{oquab2023dinov2, stein2023exposing}
following the GPIC evaluation protocol.
We use generator EMA weights for sampling and select classifier-free guidance (CFG)
separately for each model by minimizing gFD on GPIC Validation-50K, then report
results on Test-50K. Unless varied
below, sampling uses 50 Euler steps.

\paragraph{Hyperparameter selection.}
We first select the noise level $\sigma$ used to train the encoder and loss weight
$\lambda$ used to train the generator using JiT-T2I-\xlsize{} at 200K training steps. We compare five noise
levels and three mixing weights, as well as the \method{}-only
objective. The best
setting in this sweep is $\sigma=0.25$ and $\lambda=2$, which we use for our main results and ablations.
Appendix~\ref{sec:supp:loss-ablation} details the procedure and full sweep table.

\subsection{Generation Results}

\begin{wrapfigure}[20]{L}{0.36\textwidth}
\vspace{-\intextsep}
\centering
\includegraphics[width=0.90\linewidth]{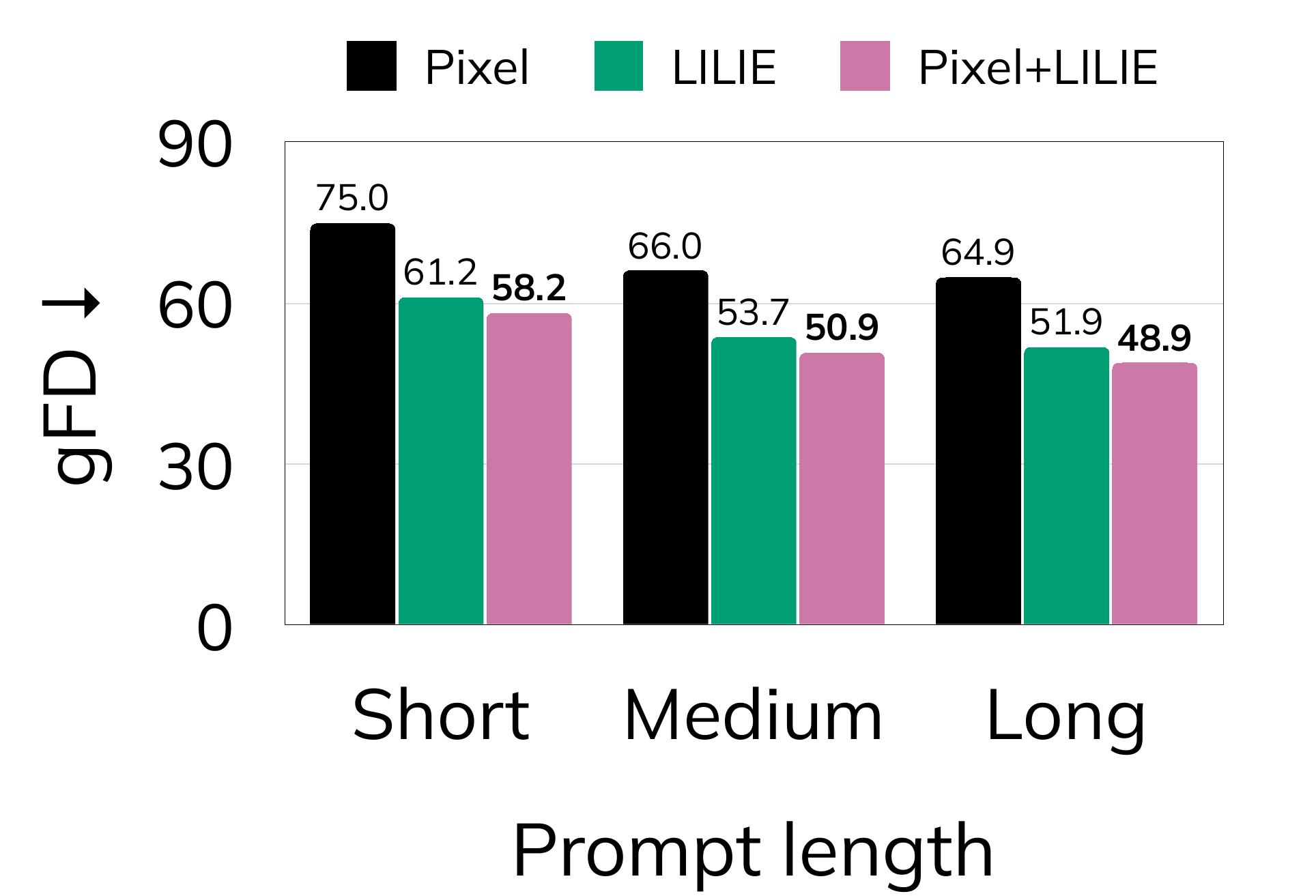}
\vspace{-3pt}
\caption{\textbf{gFD over prompt length.} We take three sets of 50K prompts from GPIC Test, each containing only short, medium, or long captions. We then sample from the Pixel, \method{}, and Pixel+\method{} JiT-T2I-\gsize{} models at their respective optimal CFG scales (Table~\ref{tab:supp:headline-results-full}). We report gFD over the GPIC Test set. Our optimal Pixel+\method{} configuration outperforms the Pixel baseline at every prompt length.}
\label{fig:main:prompt-length}
\end{wrapfigure}

\paragraph{\method{} improves generation quality.}
Adding \method{} loss to pixel loss improves gFD, precision, recall, coverage, and MMD for both generators (Table~\ref{tab:main:headline-results}). For JiT-T2I-\xlsize{}, Pixel+\method{} reduces gFD from 80.24 to 65.86, and for JiT-T2I-\gsize{}, it reduces gFD from 66.84 to 53.01. Precision and recall improve alongside gFD reduction, so our gain in generation quality results from both higher sample fidelity and better coverage of the real data distribution. We additionally report GenEval \citep{ghosh2023geneval} and DPG-Bench \citep{hu2024ella} results in Appendix~\ref{sec:supp:prompt-alignment}. Figure~\ref{fig:main:fig1} shows samples from JiT-T2I-\gsize{} trained with Pixel+\method{} loss. Figure~\ref{fig:main:prompt-length} provides gFD results when varying prompt length.

\paragraph{\method{} loss can be used on its own.}
Replacing pixel loss entirely with \method{} loss reduces gFD from 80.24 to 68.88 for JiT-T2I-\xlsize{} and from 66.84 to 55.78 for JiT-T2I-\gsize{}. In contrast, we find that perceptual loss baselines fail when used without a pixel loss term (Table~\ref{tab:main:headline-results}). Since \method{} is a bijection, matching the target image in \method{}-space exactly recovers the target in pixel space, and \method{}-space alone provides a richer training signal than pixels. \method{} therefore provides a useful training objective even without a pixel loss term, while combining the two yields the best generation quality.

\begin{wrapfigure}{r}{0.4\textwidth}
\vspace{-\intextsep}
\centering
\includegraphics[width=0.70\linewidth]{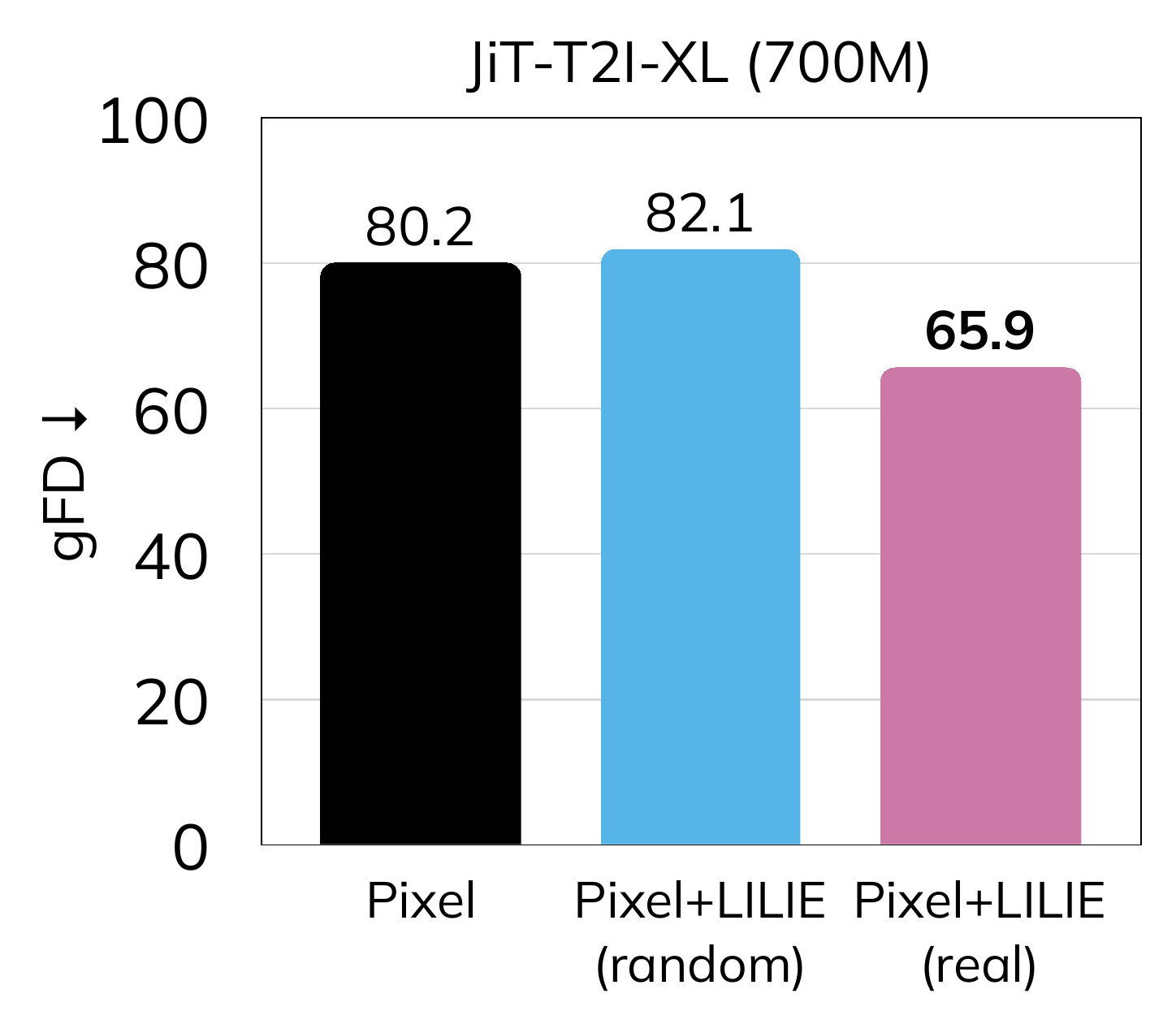}
\vspace{-13pt}
\caption{\textbf{Effect of informative captions.} gFD on GPIC Test-50K for JiT-T2I-\xlsize{} with Pixel or Pixel+\method{} loss. ``Random'' indicates replacing ground-truth captions with random token sequences during \method{} training. We observe that this prevents \method{} from learning a representation that improves generation quality.}
\label{fig:main:language-ablation}
\end{wrapfigure}

\paragraph{Language supervision matters.}
To isolate the role of informative captions, we train a \method{}-S encoder with captions comprising random token sequences. We replace each token in a real caption by an independent uniform draw from the SmolVLM-256M vocabulary, subset to tokens appearing in GPIC Full. We preserve caption lengths and VLM prompts and only randomize the encoder's training target. The downstream generator still receives the original image-caption pairs. We then train \mbox{JiT-T2I-\xlsize{}} with Pixel+\method{} loss using this encoder, using the same generator configuration as before. Figure~\ref{fig:main:language-ablation} compares this result with Pixel and Pixel+\method{} trained using real captions. With random captions, gFD is 82.06, compared with 80.24 for Pixel and 65.86 for real captions. Hence, the encoder trained on random token captions fails to improve generation quality relative to the pixel-only model, indicating that real captions play a crucial role during \method{} training.
\par\WFclear

\subsection{Sampling Steps}
\label{sec:main:sampling-steps}

\begin{figure}[!tb]
\centering
\includegraphics[width=\linewidth]{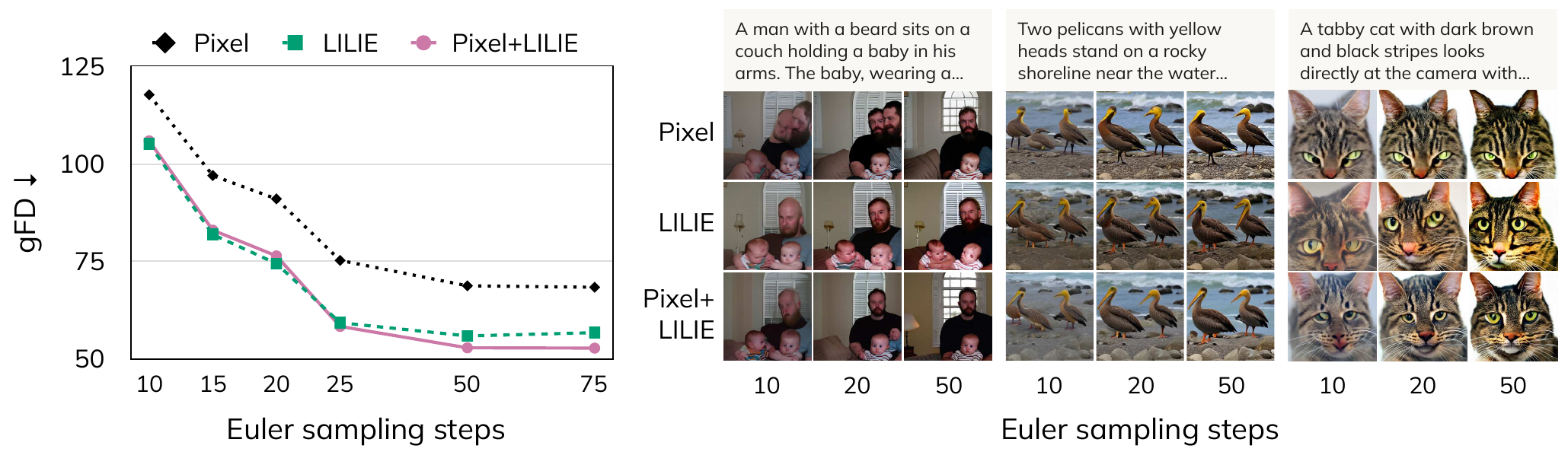}
\vspace{-6pt}
\caption{\textbf{Effect of Euler sampling steps.}
Left: JiT-T2I-\gsize{} gFD on GPIC Test-50K at 390,600 training
steps, with each model's CFG scale chosen at 50 steps and held constant.
Right: three selected qualitative examples with Euler steps increasing across columns and Pixel,
\method{}, and Pixel+\method{} down rows. Prompts and initial noise are shared
across methods and step counts.}
\label{fig:main:ablation-euler-steps}
\end{figure}

We next vary the number of Euler steps over $10,15,20,25,50,75$
(Figure~\ref{fig:main:ablation-euler-steps}). The curves
and examples compare JiT-T2I-\gsize{} models trained to 390,600 steps. Both Pixel and Pixel+\method{} benefit from increasing the budget from 10 to 75
steps. Pixel+\method{} has lower gFD than Pixel at every tested sampling budget. At 25 steps,
Pixel+\method{} reaches a gFD of 58.23, already below Pixel with 50 steps at 68.67.
The learned loss can therefore
improve the tradeoff between generation quality and the number of sampling
steps.
Beyond 50 steps, the gains are small. Moving to 75 steps reduces gFD
from 68.67 to 68.29 for Pixel and from 52.73 to 52.65 for Pixel+\method{}. \method{} sees an increase in gFD from 55.78 to 56.63 from 50 to 75 steps. We retain 50 Euler steps to compare all methods as both Pixel and Pixel+\method{} see minimal gains from moving to 75 steps, and our choice is consistent with JiT-T2I sampling in GPIC \citep{chandrasegaran2026gpic}. Figure~\ref{fig:supp:sampling-trajectories} additionally provides examples of JiT clean image predictions along a 50 step trajectory.

\subsection{Encoder Size and Data Scaling}
\label{sec:main:model-data-ablations}

\begin{figure}[!b]
\centering
\includegraphics[width=\linewidth]{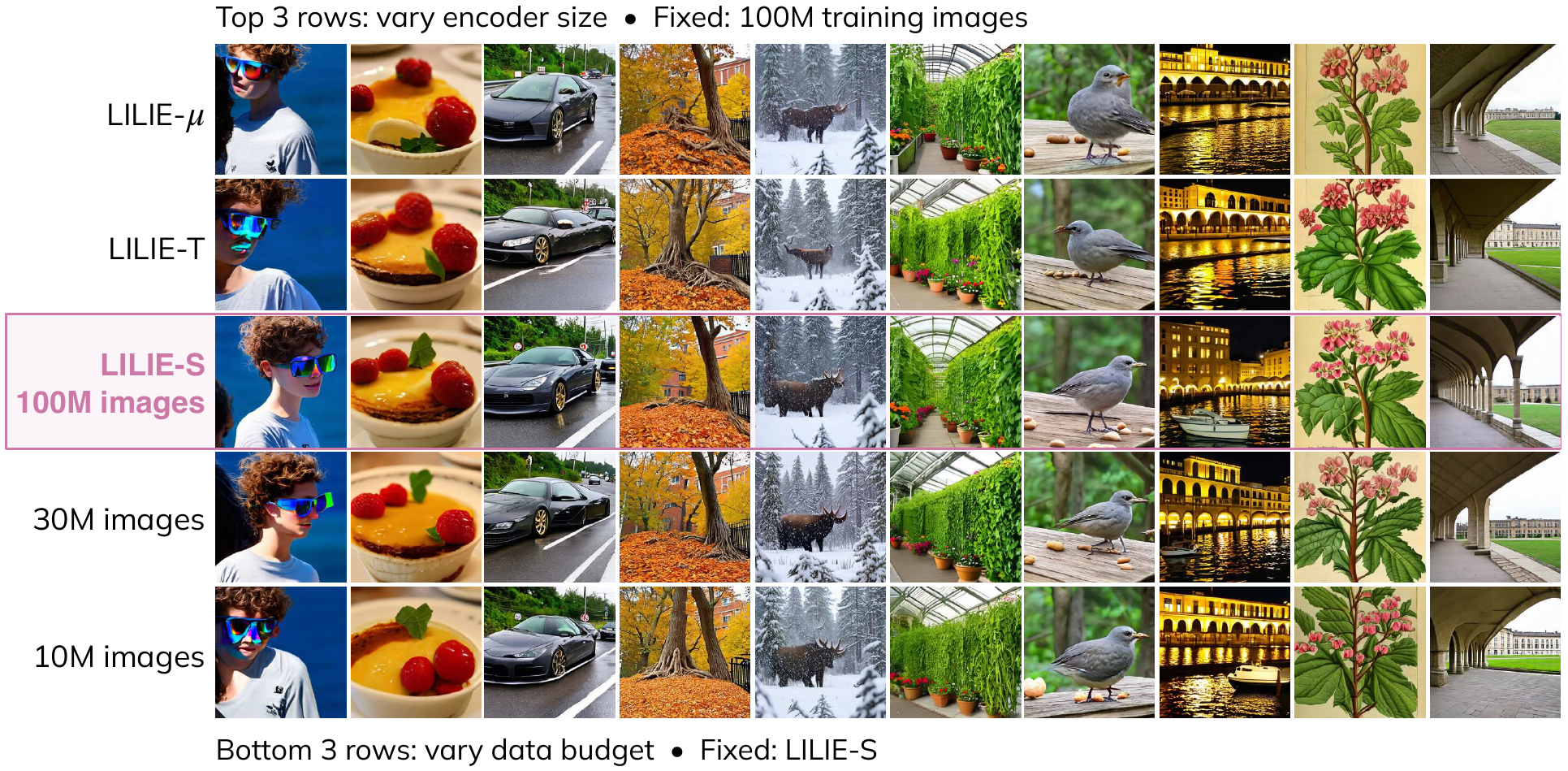}
\vspace{-10pt}
\caption{\textbf{Qualitative samples over \method{} model size and training data.}
Selected qualitative JiT-T2I-\gsize{} samples from GPIC 
use the same prompt and initial noise in each column.
The highlighted row is \method{}-S/100M, our main configuration. Model size decreases above it, and training data
decreases below it. All generators are trained for 200K steps.}
\label{fig:main:ablation-model-data}
\end{figure}

We vary the model size and training budget for \method{}
while keeping the generator setup fixed.
All generators in this ablation are trained for 200K steps. The encoder rows use Pixel+\method{} loss.
The \method{}-S encoder trained on 100M images is the common setting for both
comparisons, highlighted in Table~\ref{tab:main:ablation-model-size-fd}, Table~\ref{tab:main:ablation-data-fd},
and Figure~\ref{fig:main:ablation-model-data}. We provide gFD decompositions, PRDC, and MMD results in Appendix~\ref{sec:supp:model-data-ablations}.

\begin{table}[!t]
\centering
\small
\setlength{\tabcolsep}{5pt}
\renewcommand{\arraystretch}{1.05}
\begin{minipage}[t]{0.48\textwidth}
\centering
\caption{\textbf{Model size ablation.} gFD~$\downarrow$ at 200K generator steps. All \method{} encoders train on 100M images. The default \method{}-S encoder row is highlighted.\strut}
\label{tab:main:ablation-model-data-fd}
\label{tab:main:ablation-model-size-fd}
\setlength{\tabcolsep}{3pt}
\begin{tabular}{@{}lrrr@{}}
\toprule
Encoder & Params & JiT-T2I-\xlsize{} & JiT-T2I-\gsize{} \\
\midrule
\textcolor{black!60}{Pixel} & \textcolor{black!60}{--} & \textcolor{black!60}{114.59} & \textcolor{black!60}{101.66} \\
\midrule
\method{}-$\mu$ & 1.5M & 96.11 & 84.66 \\
\method{}-T & 9.1M & 92.66 & 80.87 \\
\rowcolor{okpurple!18}
\method{}-S & 32M & \textbf{91.79} & \textbf{76.59} \\
\bottomrule
\end{tabular}
\end{minipage}\hfill
\begin{minipage}[t]{0.48\textwidth}
\centering
\caption{\textbf{Training data ablation.} gFD~$\downarrow$ at 200K generator steps. All encoders are \method{}-S. The default 100M-image training budget row is highlighted.\strut}
\label{tab:main:ablation-data-fd}
\begin{tabular}{lrr}
\toprule
\# Training Images & JiT-T2I-\xlsize{} & JiT-T2I-\gsize{} \\
\midrule
\textcolor{black!60}{Pixel} & \textcolor{black!60}{114.59} & \textcolor{black!60}{101.66} \\
\midrule
10M & 97.79 & 83.01 \\
30M & 93.52 & 81.18 \\
\rowcolor{okpurple!16}
100M & \textbf{91.79} & \textbf{76.59} \\
\bottomrule
\end{tabular}
\end{minipage}
\end{table}

\paragraph{Model size.}
We train \method{}-$\mu$ (micro), \method{}-T (tiny), and \method{}-S (small) encoders
containing 1.5M, 9.1M, and 32M parameters, respectively. All encoders are trained for 1 epoch
on GPIC Full (100M image-caption pairs). Increasing encoder
size improves gFD for both generators. For JiT-T2I-\gsize{}, gFD decreases from 84.66
with \method{}-$\mu$ to 80.87 with \method{}-T and 76.59 with \method{}-S. Even \method{}-$\mu$ improves
over the Pixel baseline of 101.66, so a 1.5M parameter encoder already provides
a strong training signal. We detail the model configurations in Table~\ref{tab:supp:config-llsvr}.

\paragraph{Training data.}
With the \method{}-S architecture fixed, increasing \method{} training from 10M
to 30M and 100M images reduces gFD from 97.79 to 93.52 to 91.79,
respectively, for JiT-T2I-\xlsize{}. For \gsize{}, the same budgets reduce gFD from
83.01 to 81.18 and 76.59. Hence, increasing the encoder's data and training budget improves
generation quality without increasing the generator's own training budget.
\par

\section{Discussion and Conclusion}
\label{sec:main:discussion}

\begin{wrapfigure}{r}{0.48\textwidth}
\vspace{-\intextsep}
\centering
\includegraphics[width=0.75\linewidth]{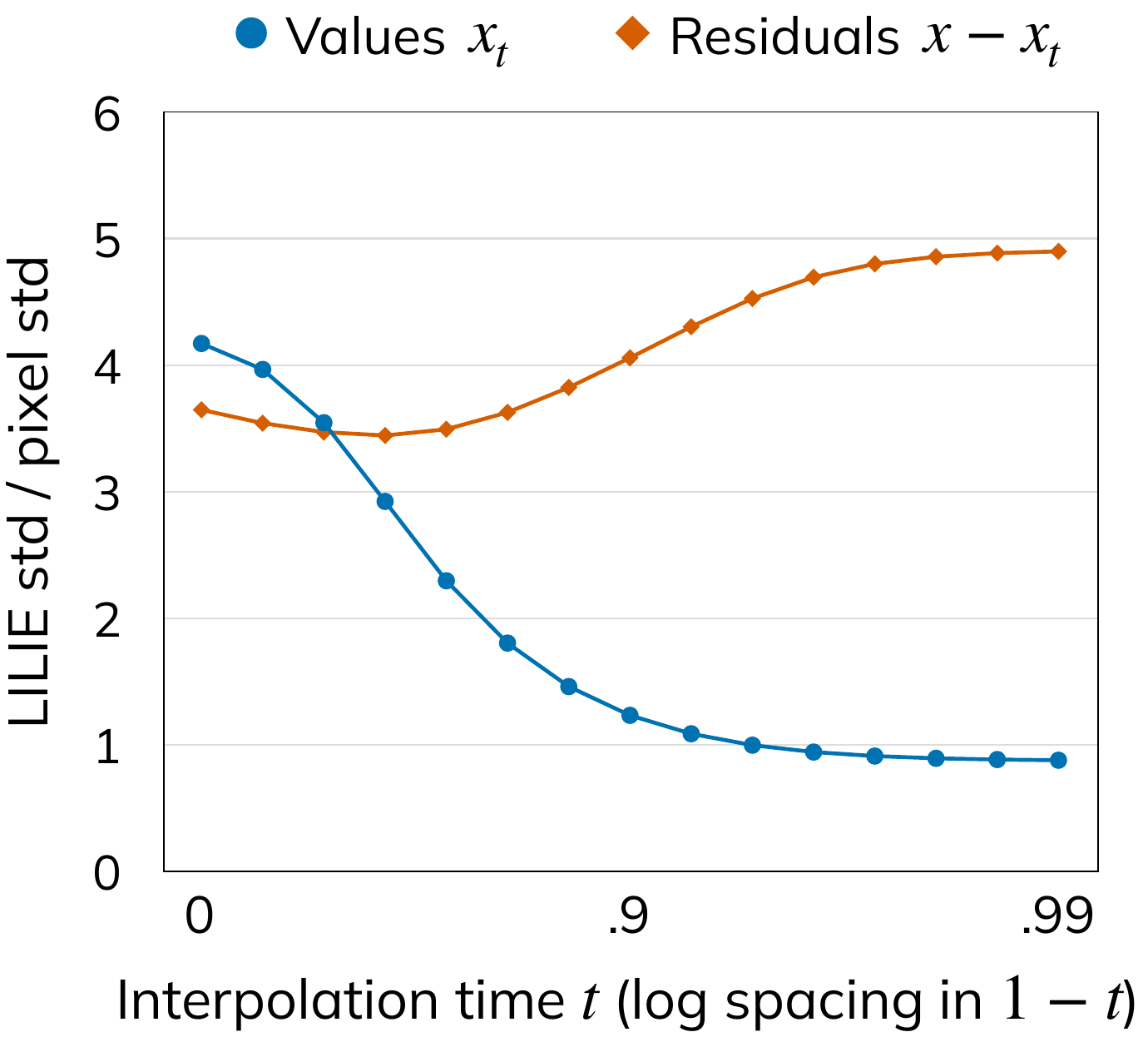}
\vspace{-7pt}
\caption{\textbf{\method{} provides a stronger training signal closer to the data distribution.}  Average standard deviation ratios for \method{} versus pixel values $x_t$, and residuals from clean images $x$, using the \method{}-S encoder on GPIC Nano (1M train images). $x_t$ denotes linear interpolation between pure noise ($t=0$) and real data \mbox{($t=1$)}.}
\label{fig:main:rich-training-signal}
\end{wrapfigure}

We have shown that language supervision alone is sufficient to learn a feature space that improves downstream generation quality. Pixel-space losses measure differences without regard to image content. \method{} instead learns its representation through language supervision. Information needed to describe an image must survive corruption, while invertibility guarantees that no visual information is lost. Figure~\ref{fig:main:rich-training-signal} illustrates an important property of the proposed language-informed loss. Near real data, at $t=0.99$, the learned representation amplifies residual standard deviation by approximately $4.9\times$ relative to pixels despite near-equal scale between raw pixel values and \method{} features. Hence \method{} makes these small deviations more prominent in the supervision and provides a stronger training signal closer to the data distribution, whereas pixel differences provide a comparatively weak signal.

\par\vspace{\baselineskip}\WFclear

\paragraph{Comparisons with pretrained perceptual losses.}
We provide results in Table~\ref{tab:main:headline-results}, and Table~\ref{tab:main:loss-comparison} summarizes the feature space we compare. \method-only loss improves generation quality over Pixel by 11 gFD points ($66.84 \to 55.78$), while DINOv2, SigLIP 2, and LPIPS losses result in degenerate quality when used without a Pixel loss term. When combined with Pixel loss, \method{}-loss achieves comparable quality to perceptual loss baselines. We recognize that prior work has had success in combining multiple feature-space losses to supervise visual generation~\citep{ma2026pixelgen}, and we leave experiments of combining Pixel, \method{}, and perceptual supervision to future work. We discuss limitations of our work in Appendix~\ref{sec:supp:limitations}.

\begin{table}[!ht]
\centering
\caption{\textbf{Comparison of feature-space losses.} ``GPIC compliant'' indicates compliance with the GPIC evaluation protocol. ``SSL'' refers to self-supervised learning, and ``Language'' refers to language-supervised learning. The VGG16 backbone is label-supervised, trained for ImageNet classification. SigLIP 2 is trained with both language-supervised and SSL objectives. Our method is only supervised by language.}
\label{tab:main:loss-comparison}
\fontsize{8}{9.5}\selectfont
\setlength{\tabcolsep}{3pt}
\begin{tabular}{@{}lcccccc@{}}
\toprule
Loss & Supervised & SSL & Language & Training images & Not distilled & GPIC compliant \\
\midrule
DINOv2 & - & \textcolor{black}{\ding{51}} & - & 142M & \xmark & \xmark \\
SigLIP 2 & - & \textcolor{black}{\ding{51}} & \textcolor{black}{\ding{51}} & 10B & \xmark & \xmark \\
LPIPS (VGG16) & \textcolor{black}{\ding{51}} & - & - & 1.28M & \cmark & \cmark \\
\rowcolor{okpurple!16}
\method{} & - & - & \textcolor{black}{\ding{51}} & 100M & \cmark & \cmark \\
\bottomrule
\end{tabular}

\end{table}

We note that DINOv2 and SigLIP 2 introduce different pretraining data and budgets. DINOv2 uses LVD-142M (142M images) \citep{oquab2023dinov2}, while SigLIP 2 trains on WebLI, a corpus of 10B images \citep{tschannen2025siglip}. Without access to the same pretraining corpora and compute, we cannot match the foundation models' data distributions.
Furthermore, we note that models trained with DINOv2 and SigLIP 2 features violate the GPIC evaluation protocol \citep{chandrasegaran2026gpic}. Both are large vision foundation models trained on sets of data that may overlap with the GPIC test set. Furthermore, GPIC evaluation uses DINOv2 features itself. We therefore regard DINOv2- and SigLIP 2-based losses as useful comparisons between pretrained representations but not as controlled experiments.

\paragraph{A simple language-informed loss for visual generation.}
In this work, we propose a simple language-informed loss for visual generation, which uses a small 32M parameter invertible encoder for improving end-to-end visual generation. With \method{}, we improve JiT-T2I-\gsize{} gFD from 66.84 to 55.78, while combining the losses reaches 53.01 (Table~\ref{tab:main:headline-results}). Our code, encoders, and generative models are available at \href{https://suchiragarwal.com/lilie}{\textcolor{okblue}{suchiragarwal.com/lilie}}.

\paragraph{Acknowledgements.}
We thank members of the Stanford Vision Lab and the CogAI group for their feedback.

\bibliography{references}
\bibliographystyle{iclr2027_conference}

\appendix
\renewcommand{\thefigure}{\thesection.\arabic{figure}}
\renewcommand{\thetable}{\thesection.\arabic{table}}
\numberwithin{figure}{section}
\numberwithin{table}{section}

\clearpage
\section*{Appendix}
\startcontents[appendix]
\printcontents[appendix]{}{1}{\setcounter{tocdepth}{2}}

\section{Additional Training Details}
\label{sec:supp:algorithm-details}

\paragraph{\texttt{preprocess} in Algorithm~\ref{alg:llsvr-training}.}
Each GPIC image is resized so that its shorter side is 256 pixels using bicubic interpolation with antialiasing, then center cropped to $256 \times 256$. Pixel values ${p \in [0, 255]}$ are mapped to $[-1, 1]$ as $2p/255 - 1$.

\paragraph{SmolVLM input.}
The noisy features are passed to SmolVLM-256M in place of an image. We do not clamp them to $[-1, 1]$, allowing values outside the usual image range. Since the VLM is finetuned with LoRA during \method{} training, it adapts to these previously out-of-domain input values.

\subsection{Encoder Architecture}
\label{sec:supp:encoder-architecture}

\begin{figure}[!htb]
    \centering
    \includegraphics[width=0.95\linewidth]{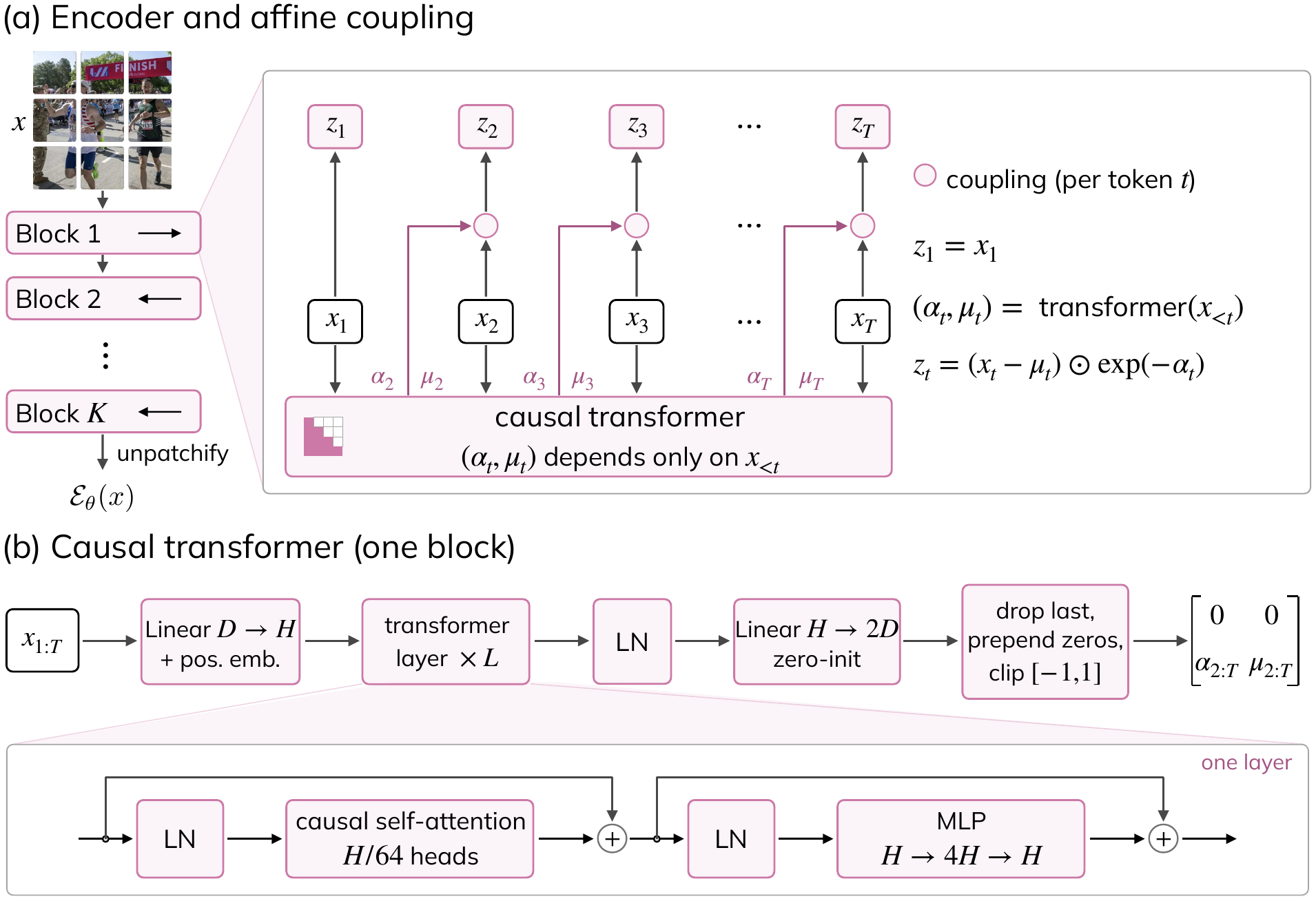}
    \caption{\textbf{Encoder implementation details.} (a) The block stack and
    autoregressive affine map. The tuple $\gamma_t=(\alpha_t,\mu_t)$ contains the
    log-scale and shift of patch $t$. (b) The causal transformer within one block,
    with bottleneck dimension $H$ and $L$ layers. LN denotes layer normalization.
    The lower row expands one transformer layer.}
    \label{fig:supp:llsvr-encoder-details}
\end{figure}

Figure~\ref{fig:supp:llsvr-encoder-details} expands the encoder in
Figure~\ref{fig:main:llsvr-training}. Each image is split into $T$ patches of dimension
$D$. The encoder applies $K$ autoregressive affine blocks, alternating the patch
order used by successive blocks, and reassembles the outputs into an image-shaped
feature tensor.

\paragraph{Block parameters.}
For patch $t$, the coupling parameters
$(\alpha_t,\mu_t)$ are log-scale and shift vectors. They depend only on patches
preceding position $t$ in the block's order. Both vectors are clipped to $[-1,1]$
before applying the affine map shown in Figure~\ref{fig:supp:llsvr-encoder-details}(a).
The first patch passes through unchanged.

\paragraph{Transformer internals.}
Each block projects patch vectors from $D$ dimensions to bottleneck dimension
$H$, adds learned positional embeddings, and applies $L$ causal transformer
layers. Each layer uses $H/64$ attention heads and an MLP with expansion ratio
four and GELU activation, with residual connections around attention and the MLP.
Attention uses layer normalization. The final zero-initialized linear head produces $2D$ values per position. We drop the last
position, prepend zeros, and clip the result. This shift makes each patch's parameters depend only on earlier
patches. The model-size ablation varies $K$, $H$, and $L$, and the configurations are detailed in Table~\ref{tab:supp:config-llsvr}.

\newpage
\section{Experimental Details}
\label{sec:supp:experimental-details}

\begin{figure}[!htbp]
  \centering
  \includegraphics[width=0.8\linewidth]{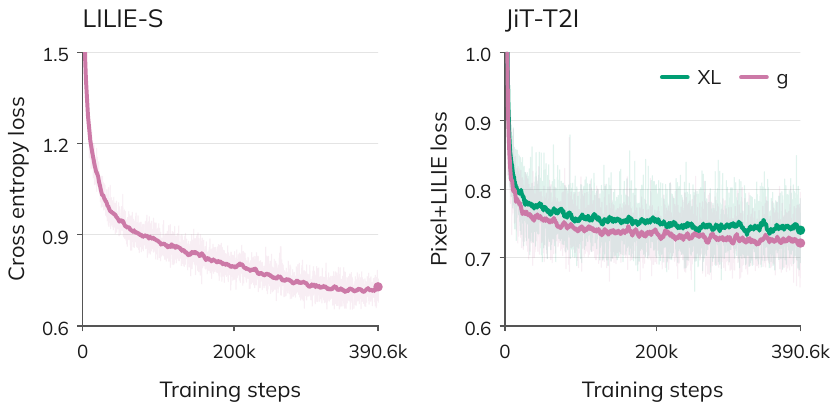}
  \caption{Left: Training loss for LILIE-S over one epoch on GPIC Full. Right: Training loss for JiT-T2I-XL and -g trained with Pixel+LILIE loss ($\lambda=2$) over one epoch on GPIC Full.}
  \label{fig:supp:training-losses}
\end{figure}

\begin{figure}[!htb]
\centering
\includegraphics[width=\linewidth]{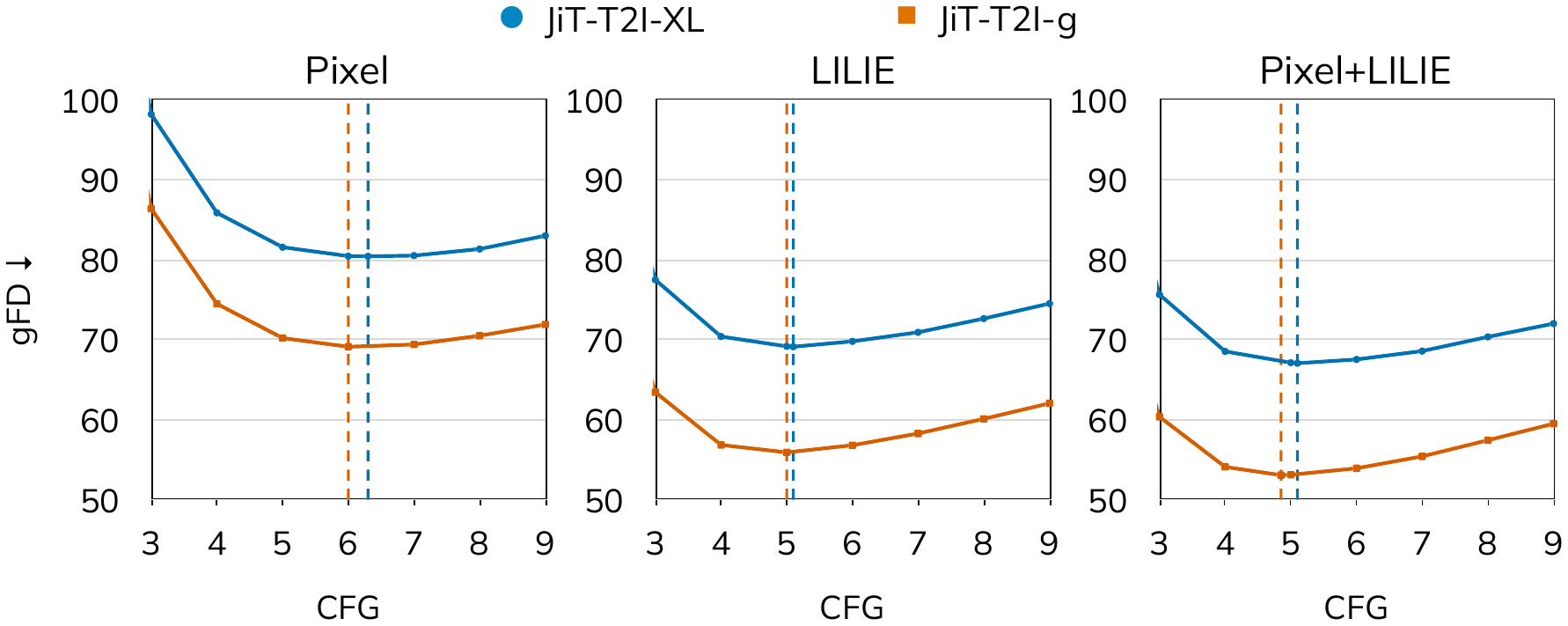}
\caption{\textbf{gFD over CFG on GPIC Validation-50K.}
We show gFD on the GPIC Validation-50K set against CFG for JiT-T2I-\xlsize{}
and JiT-T2I-\gsize{} models for Pixel, \method{}, and Pixel+\method{}.
Selected CFG scales in Table~\ref{tab:supp:headline-results-full} are indicated with dashed
vertical lines.}
\label{fig:supp:cfg-sweep}
\end{figure}

Table~\ref{tab:supp:config-llsvr} and Table~\ref{tab:supp:config-jit} summarize configurations for \method{} and JiT training. Table~\ref{tab:supp:headline-results-full} expands on Table~\ref{tab:main:headline-results}, including gFD decomposition and selected CFG values. Figure~\ref{fig:supp:training-losses} shows training loss curves for \method{}-S and JiT-T2I-\xlsize{} and -\gsize{} models trained with Pixel+\method{} loss. Figure~\ref{fig:supp:cfg-sweep} shows gFD on GPIC Validation-50K over CFG
for JiT-T2I-\xlsize{} and JiT-T2I-\gsize{} trained with Pixel, \method{}, and Pixel+\method{}. Unless otherwise specified, all results are at 390,600 training steps.

\begin{table}[htb]
\caption{\textbf{Full generation results on GPIC Test-50K.} Models train for one epoch on GPIC-Full (390,600 steps), and we train selected configurations for two epochs (781,200 steps). All results use 50 Euler steps. CFG is selected on GPIC Validation-50K at 0.05 granularity. For JiT-T2I-\gsize{}, LPIPS, DINOv2, and SigLIP 2 denote perceptual losses in
the corresponding feature space, used alone and combined with the pixel loss. Despite the large compute cost associated with each pretraining run (Section~\ref{sec:main:experiments}), we perform three one-epoch pretraining trials for Pixel and Pixel+\method{} at both generator sizes (\method{} seeds \texttt{100,101,102}, JiT-T2I seeds \texttt{0,1,2}). For these one-epoch configurations, we report mean $\pm$ one sample standard deviation for each metric over the three trials. gFD is decomposed into mean ($\mathrm{gFD}_{\mu}$) and
covariance ($\mathrm{gFD}_{\Sigma}$) components. Precision, recall, density,
and coverage measure sample fidelity and diversity. MMD is reported in raw squared units.}
\label{tab:supp:headline-results-full}
\centering
\small
\setlength{\tabcolsep}{4pt}
\begin{adjustbox}{max width=\linewidth}
\begin{tabular}{llccccccccc}
\toprule
Generator & Loss & CFG & gFD $\downarrow$ & $\mathrm{gFD}_{\mu}\downarrow$ & $\mathrm{gFD}_{\Sigma}\downarrow$ & Precision $\uparrow$ & Recall $\uparrow$ & Density $\to 1$ & Coverage $\uparrow$ & MMD $\downarrow$ \\
\midrule
\multirow{3}{*}{\shortstack{JiT-T2I-\xlsize{}\\(700M)}} & Pixel & 6.30 & 80.24{\scriptsize$\,\pm\,$1.01} & 14.26{\scriptsize$\,\pm\,$0.55} & 65.97{\scriptsize$\,\pm\,$0.46} & 0.932{\scriptsize$\,\pm\,$0.003} & 0.793{\scriptsize$\,\pm\,$0.003} & 0.979{\scriptsize$\,\pm\,$0.005} & 0.913{\scriptsize$\,\pm\,$0.003} & 0.052{\scriptsize$\,\pm\,$0.002} \\
 & \cellcolor{okpurple!16}\method{} & \cellcolor{okpurple!16}5.10 & \cellcolor{okpurple!16}68.88 & \cellcolor{okpurple!16}12.59 & \cellcolor{okpurple!16}56.28 & \cellcolor{okpurple!16}0.941 & \cellcolor{okpurple!16}0.829 & \cellcolor{okpurple!16}0.996 & \cellcolor{okpurple!16}0.930 & \cellcolor{okpurple!16}0.046 \\
 & \cellcolor{okpurple!16}Pixel+\method{} & \cellcolor{okpurple!16}5.10 & \cellcolor{okpurple!16}\textbf{65.86}{\scriptsize$\,\pm\,$0.86} & \cellcolor{okpurple!16}\textbf{11.65}{\scriptsize$\,\pm\,$0.30} & \cellcolor{okpurple!16}\textbf{54.20}{\scriptsize$\,\pm\,$0.57} & \cellcolor{okpurple!16}\textbf{0.943}{\scriptsize$\,\pm\,$0.001} & \cellcolor{okpurple!16}\textbf{0.832}{\scriptsize$\,\pm\,$0.003} & \cellcolor{okpurple!16}\textbf{1.005}{\scriptsize$\,\pm\,$0.008} & \cellcolor{okpurple!16}\textbf{0.932}{\scriptsize$\,\pm\,$0.002} & \cellcolor{okpurple!16}\textbf{0.042}{\scriptsize$\,\pm\,$0.001} \\
\midrule
\multirow{9}{*}{\shortstack{JiT-T2I-\gsize{}\\(1.1B)}} & Pixel & 6.00 & 66.84{\scriptsize$\,\pm\,$1.72} & 10.20{\scriptsize$\,\pm\,$0.66} & 56.64{\scriptsize$\,\pm\,$1.07} & 0.940{\scriptsize$\,\pm\,$0.002} & 0.825{\scriptsize$\,\pm\,$0.002} & 1.000{\scriptsize$\,\pm\,$0.016} & 0.926{\scriptsize$\,\pm\,$0.002} & 0.038{\scriptsize$\,\pm\,$0.002} \\
\cmidrule(lr){2-11}
 & DINOv2 & 7.20 & 3124.01 & 1556.14 & 1567.87 & 0.987 & 0.000 & 1.082 & 0.048 & 15.225 \\
 & SigLIP 2 & 13.50 & 3161.15 & 1535.85 & 1625.30 & 0.952 & 0.000 & 1.172 & 0.050 & 14.817 \\
 & LPIPS (VGG16) & 6.15 & 158.45 & 45.97 & 112.48 & 0.922 & 0.651 & 1.053 & 0.848 & 0.174 \\
 & \cellcolor{okpurple!16}\method{} & \cellcolor{okpurple!16}5.00 & \cellcolor{okpurple!16}55.78 & \cellcolor{okpurple!16}8.48 & \cellcolor{okpurple!16}47.30 & \cellcolor{okpurple!16}0.947 & \cellcolor{okpurple!16}0.855 & \cellcolor{okpurple!16}1.028 & \cellcolor{okpurple!16}0.942 & \cellcolor{okpurple!16}0.031 \\
\cmidrule(lr){2-11}
 & Pixel+DINOv2 & 4.60 & \textbf{47.88} & \textbf{6.09} & \textbf{41.79} & \textbf{0.951} & \textbf{0.864} & \textbf{1.051} & \textbf{0.950} & \textbf{0.023} \\
 & Pixel+SigLIP 2 & 4.60 & 52.55 & 7.03 & 45.52 & 0.956 & 0.847 & 1.064 & 0.949 & 0.026 \\
 & Pixel+LPIPS & 5.15 & 48.65 & 5.90 & 42.74 & 0.950 & 0.858 & 1.035 & 0.943 & 0.022 \\
 & \cellcolor{okpurple!16}Pixel+\method{} & \cellcolor{okpurple!16}4.85 & \cellcolor{okpurple!16}53.01{\scriptsize$\,\pm\,$0.27} & \cellcolor{okpurple!16}7.43{\scriptsize$\,\pm\,$0.05} & \cellcolor{okpurple!16}45.58{\scriptsize$\,\pm\,$0.32} & \cellcolor{okpurple!16}0.947{\scriptsize$\,\pm\,$0.001} & \cellcolor{okpurple!16}0.857{\scriptsize$\,\pm\,$0.001} & \cellcolor{okpurple!16}1.029{\scriptsize$\,\pm\,$0.007} & \cellcolor{okpurple!16}0.943{\scriptsize$\,\pm\,$0.003} & \cellcolor{okpurple!16}0.028{\scriptsize$\,\pm\,$0.0001} \\
\midrule
\multicolumn{11}{l}{\textit{Two epochs (781,200 training steps)}} \\
\midrule
\multirow{5}{*}{\shortstack{JiT-T2I-\gsize{}\\(1.1B)}} & Pixel & 5.00 & 49.20 & 6.79 & 42.41 & 0.951 & 0.871 & 1.025 & 0.943 & 0.025 \\
 & Pixel+DINOv2 & 3.85 & \textbf{38.29} & \textbf{4.86} & \textbf{33.43} & \textbf{0.959} & \textbf{0.883} & \textbf{1.069} & \textbf{0.957} & \textbf{0.018} \\
 & Pixel+SigLIP 2 & 3.90 & 40.27 & 5.15 & 35.12 & 0.956 & 0.881 & 1.066 & 0.959 & 0.019 \\
 & Pixel+LPIPS & 4.50 & 39.35 & 4.51 & 34.84 & 0.952 & 0.882 & 1.043 & 0.953 & 0.017 \\
 & \cellcolor{okpurple!16}Pixel+\method{} & \cellcolor{okpurple!16}3.90 & \cellcolor{okpurple!16}40.18 & \cellcolor{okpurple!16}5.37 & \cellcolor{okpurple!16}34.82 & \cellcolor{okpurple!16}0.954 & \cellcolor{okpurple!16}0.886 & \cellcolor{okpurple!16}1.058 & \cellcolor{okpurple!16}0.959 & \cellcolor{okpurple!16}0.020 \\
\bottomrule
\end{tabular}

\end{adjustbox}
\end{table}

\begin{table}[htb]
\caption{\textbf{\method{} configurations.}}
\label{tab:supp:config-llsvr}
\centering
\tablestyle{4pt}{1.05}
\begin{tabular}{l | x{70}x{70}x{70}}
 & \textbf{\method{}-$\mu$} & \textbf{\method{}-T} & \textbf{\method{}-S} \\
\shline
\rowcolor[gray]{0.9}\multicolumn{4}{l}{\textbf{architecture}} \\
flow backbone & \multicolumn{3}{c}{Improved TARFlow \citep{zhai2024tarflow,lu2025bidirectional}} \\
blocks $K$ & 2 & 4 & 4 \\
layers per block $L$ & 2 & 4 & 4 \\
bottleneck dim $H$ & 128 & 192 & 384 \\
heads & 2 & 3 & 6 \\
parameters & 1.5M & 9.1M & 32M \\
image size & \multicolumn{3}{c}{256} \\
patch size & \multicolumn{3}{c}{16} \\
coupling clip range & \multicolumn{3}{c}{1.0} \\
VLM & \multicolumn{3}{c}{SmolVLM-256M \citep{marafioti2025smolvlm}} \\
LoRA \citep{hu2022lora} rank & \multicolumn{3}{c}{64} \\
LoRA $\alpha$ & \multicolumn{3}{c}{32} \\
LoRA targets & \multicolumn{3}{c}{all linear layers except the output head} \\
max caption tokens & \multicolumn{3}{c}{256} \\
\midline
\rowcolor[gray]{0.9}\multicolumn{4}{l}{\textbf{training}} \\
training steps & \multicolumn{3}{c}{390,600 (1 epoch on GPIC Full)} \\
warmup steps & \multicolumn{3}{c}{20,000} \\
optimizer & \multicolumn{3}{c}{AdamW \citep{loshchilov2019decoupled}, $\beta_1, \beta_2 = 0.9, 0.95$} \\
batch size & \multicolumn{3}{c}{256} \\
flow learning rate & 1.2e-3 & 8e-4 & 4e-4 \\
VLM learning rate & \multicolumn{3}{c}{4e-4} \\
learning rate schedule & \multicolumn{3}{c}{linear warmup, cosine decay to 1e-6} \\
weight decay & \multicolumn{3}{c}{1e-4} \\
gradient clipping & \multicolumn{3}{c}{AutoClip \citep{seetharaman2020autoclip}, 10th percentile} \\
noise level $\sigma$ & \multicolumn{3}{c}{0.25} \\
feature RMS target & \multicolumn{3}{c}{1/2} \\
ema decay for $m_2$ & \multicolumn{3}{c}{0.99} \\
\end{tabular}
\end{table}

\clearpage

\begin{table}[!h]
\caption{\textbf{Generative model configurations.}}
\label{tab:supp:config-jit}
\centering
\tablestyle{4pt}{1.05}
\begin{tabular}{l | x{105}x{105}}
 & \textbf{JiT-T2I-\xlsize{}} & \textbf{JiT-T2I-\gsize{}} \\
\shline
\rowcolor[gray]{0.9}\multicolumn{3}{l}{\textbf{architecture}} \\
depth & 24 & 16 \\
hidden dim & 1024 & 1536 \\
heads & 16 & 24 \\
bottleneck & 128 & 256 \\
parameters & 700M & 1.1B \\
image size & \multicolumn{2}{c}{256} \\
patch size & \multicolumn{2}{c}{16} \\
text encoder & \multicolumn{2}{c}{Qwen3-1.7B \citep{qwen3}, frozen} \\
max caption tokens & \multicolumn{2}{c}{300} \\
\midline
\rowcolor[gray]{0.9}\multicolumn{3}{l}{\textbf{training}} \\
training steps & \multicolumn{2}{c}{200K (ablations), 390,600, 781,200} \\
warmup steps & \multicolumn{2}{c}{4,000} \\
optimizer & \multicolumn{2}{c}{AdamW \citep{loshchilov2019decoupled}, $\beta_1, \beta_2 = 0.9, 0.95$} \\
batch size & \multicolumn{2}{c}{256} \\
learning rate & \multicolumn{2}{c}{1e-4} \\
learning rate schedule & \multicolumn{2}{c}{constant} \\
weight decay & \multicolumn{2}{c}{0} \\
gradient clipping & \multicolumn{2}{c}{1.0} \\
ema decay & \multicolumn{2}{c}{0.9999} \\
time sampler & \multicolumn{2}{c}{$\text{logit}(t) \sim \mathcal N(\mu, \sigma^2)$, $\mu = -0.8$, $\sigma = 0.8$} \\
noise scale & \multicolumn{2}{c}{1.0} \\
clip of $(1 - t)$ in division & \multicolumn{2}{c}{0.05} \\
caption drop (for CFG) & \multicolumn{2}{c}{0.1} \\
\midline
\rowcolor[gray]{0.9}\multicolumn{3}{l}{\textbf{sampling}} \\
ODE solver & \multicolumn{2}{c}{Euler} \\
ODE steps & \multicolumn{2}{c}{50} \\
time steps & \multicolumn{2}{c}{linear in [0.0, 1.0]} \\
CFG scale \citep{ho2022classifier} & \multicolumn{2}{c}{selected on GPIC Validation-50K} \\
weights & \multicolumn{2}{c}{EMA} \\
\end{tabular}
\end{table}

\subsection{Additional Benchmarks}
\label{sec:supp:prompt-alignment}
Tables~\ref{tab:supp:geneval} and~\ref{tab:supp:dpg-bench} report GenEval \citep{ghosh2023geneval} and DPG-Bench \citep{hu2024ella} results for the JiT-T2I-\gsize{} models (390,600 steps). We use 50 Euler steps and CFG values listed in Table~\ref{tab:supp:headline-results-full}.

\begin{table}[!ht]
\caption{GenEval scores (\%, higher is better) using the original prompts with the leading \texttt{a photo of} prefix removed, which improves scores for all 3 configurations. Errors are $\pm$ one standard error over prompts. Bold indicates the best result per column.}
\label{tab:supp:geneval}
\centering
\small
\setlength{\tabcolsep}{2pt}
\begin{tabular}{@{}lrrrrrrr@{}}
\toprule
Loss & Overall & Single & Two & Counting & Colors & Position & Color attr. \\
\midrule
Pixel & 33.57{\scriptsize$\,\pm\,$1.38} & 67.50{\scriptsize$\,\pm\,$4.34} & \textbf{28.28}{\scriptsize$\,\pm\,$3.26} & 14.69{\scriptsize$\,\pm\,$2.42} & 53.72{\scriptsize$\,\pm\,$4.21} & 18.00{\scriptsize$\,\pm\,$2.78} & 19.25{\scriptsize$\,\pm\,$2.79} \\
\cellcolor{okpurple!16}\method{} & \cellcolor{okpurple!16}32.49{\scriptsize$\,\pm\,$1.32} & \cellcolor{okpurple!16}62.50{\scriptsize$\,\pm\,$4.00} & \cellcolor{okpurple!16}26.52{\scriptsize$\,\pm\,$3.17} & \cellcolor{okpurple!16}14.06{\scriptsize$\,\pm\,$2.63} & \cellcolor{okpurple!16}\textbf{56.12}{\scriptsize$\,\pm\,$4.03} & \cellcolor{okpurple!16}13.00{\scriptsize$\,\pm\,$2.32} & \cellcolor{okpurple!16}\textbf{22.75}{\scriptsize$\,\pm\,$2.80} \\
\cellcolor{okpurple!16}Pixel+\method{} & \cellcolor{okpurple!16}\textbf{36.41}{\scriptsize$\,\pm\,$1.32} & \cellcolor{okpurple!16}\textbf{77.50}{\scriptsize$\,\pm\,$3.57} & \cellcolor{okpurple!16}24.24{\scriptsize$\,\pm\,$3.16} & \cellcolor{okpurple!16}\textbf{15.62}{\scriptsize$\,\pm\,$2.68} & \cellcolor{okpurple!16}\textbf{56.12}{\scriptsize$\,\pm\,$3.89} & \cellcolor{okpurple!16}\textbf{23.75}{\scriptsize$\,\pm\,$3.02} & \cellcolor{okpurple!16}21.25{\scriptsize$\,\pm\,$2.92} \\
\bottomrule
\end{tabular}
\end{table}

\begin{table}[!ht]
\caption{DPG-Bench scores (\%, higher is better). Errors are $\pm$ one standard error over prompts. Bold indicates the best result per column.}
\label{tab:supp:dpg-bench}
\centering
\small
\setlength{\tabcolsep}{3pt}
\begin{tabular}{@{}lrrrrrr@{}}
\toprule
Loss & Overall & Global & Entity & Attribute & Relation & Other \\
\midrule
Pixel & 73.00{\scriptsize$\,\pm\,$0.63} & 76.60{\scriptsize$\,\pm\,$2.31} & 82.21{\scriptsize$\,\pm\,$0.56} & \textbf{86.59}{\scriptsize$\,\pm\,$0.53} & 92.30{\scriptsize$\,\pm\,$0.58} & \textbf{70.00}{\scriptsize$\,\pm\,$2.91} \\
\cellcolor{okpurple!16}\method{} & \cellcolor{okpurple!16}72.86{\scriptsize$\,\pm\,$0.61} & \cellcolor{okpurple!16}76.90{\scriptsize$\,\pm\,$2.29} & \cellcolor{okpurple!16}82.13{\scriptsize$\,\pm\,$0.58} & \cellcolor{okpurple!16}86.25{\scriptsize$\,\pm\,$0.54} & \cellcolor{okpurple!16}91.68{\scriptsize$\,\pm\,$0.60} & \cellcolor{okpurple!16}64.40{\scriptsize$\,\pm\,$3.07} \\
\cellcolor{okpurple!16}Pixel+\method{} & \cellcolor{okpurple!16}\textbf{73.63}{\scriptsize$\,\pm\,$0.61} & \cellcolor{okpurple!16}\textbf{79.33}{\scriptsize$\,\pm\,$2.15} & \cellcolor{okpurple!16}\textbf{82.49}{\scriptsize$\,\pm\,$0.57} & \cellcolor{okpurple!16}86.15{\scriptsize$\,\pm\,$0.53} & \cellcolor{okpurple!16}\textbf{92.50}{\scriptsize$\,\pm\,$0.57} & \cellcolor{okpurple!16}68.80{\scriptsize$\,\pm\,$2.89} \\
\bottomrule
\end{tabular}
\end{table}

\subsection{Implementation of Perceptual Loss Baselines}
\label{sec:supp:baselines}
We follow PixelGen \citep{ma2026pixelgen} in implementing our LPIPS, DINOv2, and SigLIP 2 perceptual loss baselines. Each baseline uses only one perceptual loss, and we disable REPA feature alignment. We train JiT-T2I-\gsize{} models from scratch with the configuration from Table~\ref{tab:supp:config-jit}.

\paragraph{LPIPS.} We use the standard pretrained VGG16 LPIPS metric \citep{zhang2018perceptual}.

\paragraph{DINOv2 and SigLIP 2.} 
PixelGen uses a P-DINO perceptual loss using the DINOv2 ViT-B/14 backbone, and we follow the same methodology to construct P-SigLIP with a SigLIP 2 ViT-B/16@256px backbone. For a predicted $\hat x$ and target $x$, we compare these images in DINOv2 and SigLIP 2 feature spaces. Let $f_p$ denote the transformation from an input image in pixel space to its $p$th patch token. For SigLIP 2, these are the features prior to the final pooling head in the vision encoder. For both the DINOv2 and SigLIP 2 backbones, we have $P=256$ tokens total and $D=768$ dimensionality of each token. We then compare $f_p(\hat x)$ and $f_p(x)$ with one of the following distance functions.
\begin{align*}
    d_{\cos} (\hat x, x) &= \frac{1}{P} \sum_{p=1}^P \left( 1 - \frac{f_p (\hat x)^\top f_p(x)}{\left\lVert f_p(\hat x) \right\rVert_2 \left\lVert f_p(x) \right\rVert_2} \right),\\
    d_2(\hat x, x) &= \frac{1}{PD} \sum_{p=1}^P \left\Vert f_p(\hat x) - f_p(x) \right\Vert_2^2.
\end{align*}
Following PixelGen, we use cosine dissimilarity when combining either feature loss with the pixel loss. For feature-space-only training, we use $d_2$.

\paragraph{Aggregate form and time gating.}
For pixel + perceptual loss baselines, the loss function takes the form
\[
\mathcal L = \mathcal L_\text{Pixel} + \1_{t \geq \tau} \lambda \mathcal L_\text{Perceptual},
\]
where $\1_{t \geq \tau}$ is a simple time-based gate, turning off perceptual supervision in the high-noise regime. We use PixelGen's threshold of $\tau = 0.3$. For LPIPS, $\lambda = 0.1$, and for DINOv2 and SigLIP 2, $\lambda = 0.01$. When we train solely with the perceptual loss and without any pixel loss term, we set $\lambda = 1$ and omit the $\mathcal L_\text{Pixel}$ term and time gate entirely.

\section{Noise Level and Loss Weight}
\label{sec:supp:loss-ablation}

We select the corruption level used to train \method{} and the weight of
the learned loss using GPIC Validation-50K for evaluation. We train \method{}-S encoders
with $\sigma\in\{0,0.25,0.5,0.75,1\}$. For each encoder, we train
JiT-T2I-\xlsize{} generators with mixing weights $\lambda\in\{0.5,1,2\}$ and with
the \method{} objective. We compare generators at 200K training steps.
Each model uses 50 Euler steps, and its CFG scale is selected from the grid $\{5.50,6.25,7.00,7.75,8.50\}$ using gFD on 50K validation samples. The selected scales appear in
parentheses in Table~\ref{tab:supp:ablation-sigma-lambda}.

Among the mixed objectives, the setting $\sigma=0.25$ and $\lambda=2$ gives the lowest
validation gFD. We use this as the default setting for our 1-epoch GPIC Full training runs. We notice that encoders trained without noise result in degenerate generation quality, so corruption in the training procedure is required for learning a representation that is useful to supervise generation. Increasing noise beyond
$0.25$ does not improve the best mixed result.

\begin{table}[!htb]
\caption{\textbf{Noise level and loss weight.} gFD on GPIC Validation-50K
with JiT-T2I-\xlsize{} at 200K training steps. The selected CFG scale is in
parentheses. The Pixel baseline scores 114.90 at CFG 7.75. Bold marks
the lowest gFD in the sweep.}
\label{tab:supp:ablation-sigma-lambda}
\centering
\small
\setlength{\tabcolsep}{5pt}
\begin{adjustbox}{max width=\linewidth}
\begin{tabular}{ccccc}
\toprule
$\sigma$ & $\lambda=0.5$ & $\lambda=1$ & $\lambda=2$ & \method{}-only \\
\midrule
0 & 1832.52 (8.50) & 1854.12 (8.50) & 2165.24 (8.50) & 1502.33 (8.50) \\
0.25 & 97.12 (6.25) & 92.60 (7.00) & \textbf{92.07} (6.25) & 96.58 (6.25) \\
0.5 & 124.30 (7.75) & 125.55 (7.75) & 124.30 (7.00) & 96.72 (6.25) \\
0.75 & 118.81 (7.75) & 110.90 (7.75) & 101.66 (7.00) & 100.58 (7.00) \\
1.0 & 103.32 (7.00) & 121.37 (7.75) & 116.73 (7.00) & 106.44 (7.75) \\
\bottomrule
\end{tabular}

\end{adjustbox}
\end{table}

\section{Model Size and Data Ablations}
\label{sec:supp:model-data-ablations}

We expand on \Cref{tab:main:ablation-model-data-fd,tab:main:ablation-data-fd} to include all metrics. Table~\ref{tab:supp:ablation-flow-size} expands Table~\ref{tab:main:ablation-model-size-fd}, and Table~\ref{tab:supp:ablation-data-scaling} expands Table~\ref{tab:main:ablation-data-fd}.
All generators are evaluated at 200K training steps with 50 Euler steps. We select CFG scale via the same method described in Appendix~\ref{sec:supp:loss-ablation}. All quantitative results use GPIC Test-50K prompts. Lower is better for gFD,
its components, and MMD. Higher is better for precision, recall, and coverage. Values near 1 are ideal for density.

\begin{table}[!htbp]
\caption{\textbf{Full model-size ablation results.} The \method{} encoders are
trained on 100M images. TARFlow configuration is hidden dimension/blocks/layers (H/K/L) per block. Our primary setting is highlighted.}
\label{tab:supp:ablation-flow-size}
\centering
\footnotesize
\setlength{\tabcolsep}{3pt}
\begin{adjustbox}{max width=\linewidth}
\begin{tabular}{llccccccccccc}
\toprule
Generator & Encoder & H/K/L & Params & CFG & gFD $\downarrow$ & $\mathrm{gFD}_{\mu}\downarrow$ & $\mathrm{gFD}_{\Sigma}\downarrow$ & Precision $\uparrow$ & Recall $\uparrow$ & Density $\to 1$ & Coverage $\uparrow$ & MMD $\downarrow$ \\
\midrule
\multirow{4}{*}{JiT-T2I-\xlsize{}} & Pixel & -- & 0 & 7.75 & 114.59 & 22.07 & 92.52 & 0.924 & 0.696 & 0.940 & 0.870 & 0.082 \\
 & \method{}-$\mu$ & 128/2/2 & 1.5M & 7.00 & 96.11 & 17.98 & 78.14 & 0.931 & 0.760 & 0.957 & 0.896 & 0.066 \\
 & \method{}-T & 192/4/4 & 9.1M & 6.25 & 92.66 & 17.58 & 75.07 & 0.931 & 0.763 & 0.972 & 0.902 & 0.065 \\
 & \cellcolor{okpurple!16}\method{}-S & \cellcolor{okpurple!16}384/4/4 & \cellcolor{okpurple!16}32M & \cellcolor{okpurple!16}6.25 & \cellcolor{okpurple!16}\textbf{91.79} & \cellcolor{okpurple!16}17.39 & \cellcolor{okpurple!16}74.40 & \cellcolor{okpurple!16}0.928 & \cellcolor{okpurple!16}0.762 & \cellcolor{okpurple!16}0.973 & \cellcolor{okpurple!16}0.905 & \cellcolor{okpurple!16}0.064 \\
\midrule
\multirow{4}{*}{JiT-T2I-\gsize{}} & Pixel & -- & 0 & 7.00 & 101.66 & 18.52 & 83.14 & 0.922 & 0.737 & 0.934 & 0.880 & 0.069 \\
 & \method{}-$\mu$ & 128/2/2 & 1.5M & 6.25 & 84.66 & 14.11 & 70.55 & 0.934 & 0.788 & 0.972 & 0.915 & 0.052 \\
 & \method{}-T & 192/4/4 & 9.1M & 6.25 & 80.87 & 13.43 & 67.44 & 0.937 & 0.792 & 0.982 & 0.910 & 0.050 \\
 & \cellcolor{okpurple!16}\method{}-S & \cellcolor{okpurple!16}384/4/4 & \cellcolor{okpurple!16}32M & \cellcolor{okpurple!16}6.25 & \cellcolor{okpurple!16}\textbf{76.59} & \cellcolor{okpurple!16}12.89 & \cellcolor{okpurple!16}63.70 & \cellcolor{okpurple!16}0.941 & \cellcolor{okpurple!16}0.805 & \cellcolor{okpurple!16}0.989 & \cellcolor{okpurple!16}0.917 & \cellcolor{okpurple!16}0.047 \\
\bottomrule
\end{tabular}

\end{adjustbox}
\end{table}

\begin{table}[!htbp]
\caption{\textbf{Full training-data ablation results.} We ablate the number of images used to train the \method{}-S encoder (32M parameters). The 100M rows are the same \method{}-S configurations as in
Table~\ref{tab:supp:ablation-flow-size} and are highlighted as our primary setting.}
\label{tab:supp:ablation-data-scaling}
\centering
\footnotesize
\setlength{\tabcolsep}{3pt}
\begin{adjustbox}{max width=\linewidth}
\begin{tabular}{llccccccccc}
\toprule
Generator & \# training images & CFG & gFD $\downarrow$ & $\mathrm{gFD}_{\mu}\downarrow$ & $\mathrm{gFD}_{\Sigma}\downarrow$ & Precision $\uparrow$ & Recall $\uparrow$ & Density $\to 1$ & Coverage $\uparrow$ & MMD $\downarrow$ \\
\midrule
\multirow{3}{*}{JiT-T2I-\xlsize{}} & 10M & 7.00 & 97.79 & 18.58 & 79.22 & 0.927 & 0.751 & 0.950 & 0.895 & 0.068 \\
 & 30M & 6.25 & 93.52 & 17.67 & 75.85 & 0.930 & 0.766 & 0.969 & 0.906 & 0.065 \\
 & \cellcolor{okpurple!16}100M & \cellcolor{okpurple!16}6.25 & \cellcolor{okpurple!16}\textbf{91.79} & \cellcolor{okpurple!16}17.39 & \cellcolor{okpurple!16}74.40 & \cellcolor{okpurple!16}0.928 & \cellcolor{okpurple!16}0.762 & \cellcolor{okpurple!16}0.973 & \cellcolor{okpurple!16}0.905 & \cellcolor{okpurple!16}0.064 \\
\midrule
\multirow{3}{*}{JiT-T2I-\gsize{}} & 10M & 7.00 & 83.01 & 13.58 & 69.44 & 0.934 & 0.788 & 0.983 & 0.914 & 0.051 \\
 & 30M & 7.00 & 81.18 & 13.34 & 67.85 & 0.934 & 0.789 & 0.983 & 0.915 & 0.050 \\
 & \cellcolor{okpurple!16}100M & \cellcolor{okpurple!16}6.25 & \cellcolor{okpurple!16}\textbf{76.59} & \cellcolor{okpurple!16}12.89 & \cellcolor{okpurple!16}63.70 & \cellcolor{okpurple!16}0.941 & \cellcolor{okpurple!16}0.805 & \cellcolor{okpurple!16}0.989 & \cellcolor{okpurple!16}0.917 & \cellcolor{okpurple!16}0.047 \\
\bottomrule
\end{tabular}

\end{adjustbox}
\end{table}

\section{Limitations}
\label{sec:supp:limitations}

We perform all of our experiments on a single corpus, resolution, and generator family, namely GPIC at $256 \times 256$ with JiT-T2I. Additionally, training the encoder requires captions and a VLM, and we study only the SmolVLM-256M backbone. A larger VLM may help construct a better feature space, but it comes at an increase in encoder training cost. For \method{} training, we follow the GPIC setup of training our encoder for one epoch on GPIC Full, but we do not propose a method of determining the optimal training budget. Our largest version of \method{} is a 32M parameter encoder, and we do not experiment on scales closer to DINOv2 and SigLIP 2 base model sizes (86M parameters). We do not experiment with combinations of \method{} and perceptual losses.

\section{Captioning Prompts}
\label{sec:supp:prompts}

GPIC consists of captions in four length categories (\texttt{tag}, \texttt{short},
\texttt{medium}, \texttt{long}), and \method{} training pairs each caption type with
its own prompt for SmolVLM-256M. Figure~\ref{fig:supp:captioning-prompts}
lists the four pairs. These are condensed versions of prompts used for GPIC corpus
captioning \citep{chandrasegaran2026gpic}.

\begin{figure}[!htp]
\centering
\begin{minipage}{\linewidth}
\footnotesize
\begin{verbatim}
[tag]
system:
  Your goal is to write a compressed keyword-style caption (unordered tags)
  for the given image. The output must be a short, unordered list of tags
  describing the main visible content.
user:
  Write a keyword-style caption (tag-style) for the image shown.
\end{verbatim}
\vspace{2pt}
\begin{verbatim}
[short]
system:
  Your goal is to write a short caption (1-2 sentences) that describes the
  given image in simple language that a middle school student can
  understand. The caption must be enough to identify the main subject and
  the basic setting/action.
user:
  Write a short caption (1-2 sentences) for the image shown.
\end{verbatim}
\vspace{2pt}
\begin{verbatim}
[medium]
system:
  Your goal is to write a medium-length caption (2-3 sentences) that
  describes the given image in simple language that a middle school student
  can understand. The caption must be enough to identify the main subject
  and the basic setting/action.
user:
  Please write a caption (2-3 sentences) for the image shown.
\end{verbatim}
\vspace{2pt}
\begin{verbatim}
[long]
system:
  Your goal is to write a long, highly detailed caption (5-7 sentences)
  that describes the given image in simple language that a middle school
  student can understand. The caption should include enough detail for them
  to understand what appears in the foreground and background and picture
  how the scene is arranged.
user:
  Please write a caption (5-7 sentences) for the image shown.
\end{verbatim}
\end{minipage}
\caption{Prompts given to SmolVLM-256M during \method{} training for the four GPIC caption-length categories: \texttt{tag}, \texttt{short}, \texttt{medium}, and \texttt{long}.}
\label{fig:supp:captioning-prompts}
\label{fig:supp:prompt}
\label{fig:supp:prompt-tag}
\label{fig:supp:prompt-short}
\label{fig:supp:prompt-medium}
\label{fig:supp:prompt-long}
\end{figure}

\clearpage
\section{Samples}
\label{sec:supp:samples}

\newcommand{\uncuratedsamplescaption}{Random samples from the held-out GPIC Test-50K. Each example shows the real image from the GPIC corpus on the left, followed by the three generated samples from JiT-T2I-\gsize{} models. The first generated sample is from the Pixel baseline, and the following two are from \method{} and Pixel+\method{}.}

\begin{figure}[!h]
\centering
\includegraphics[width=0.55\linewidth]{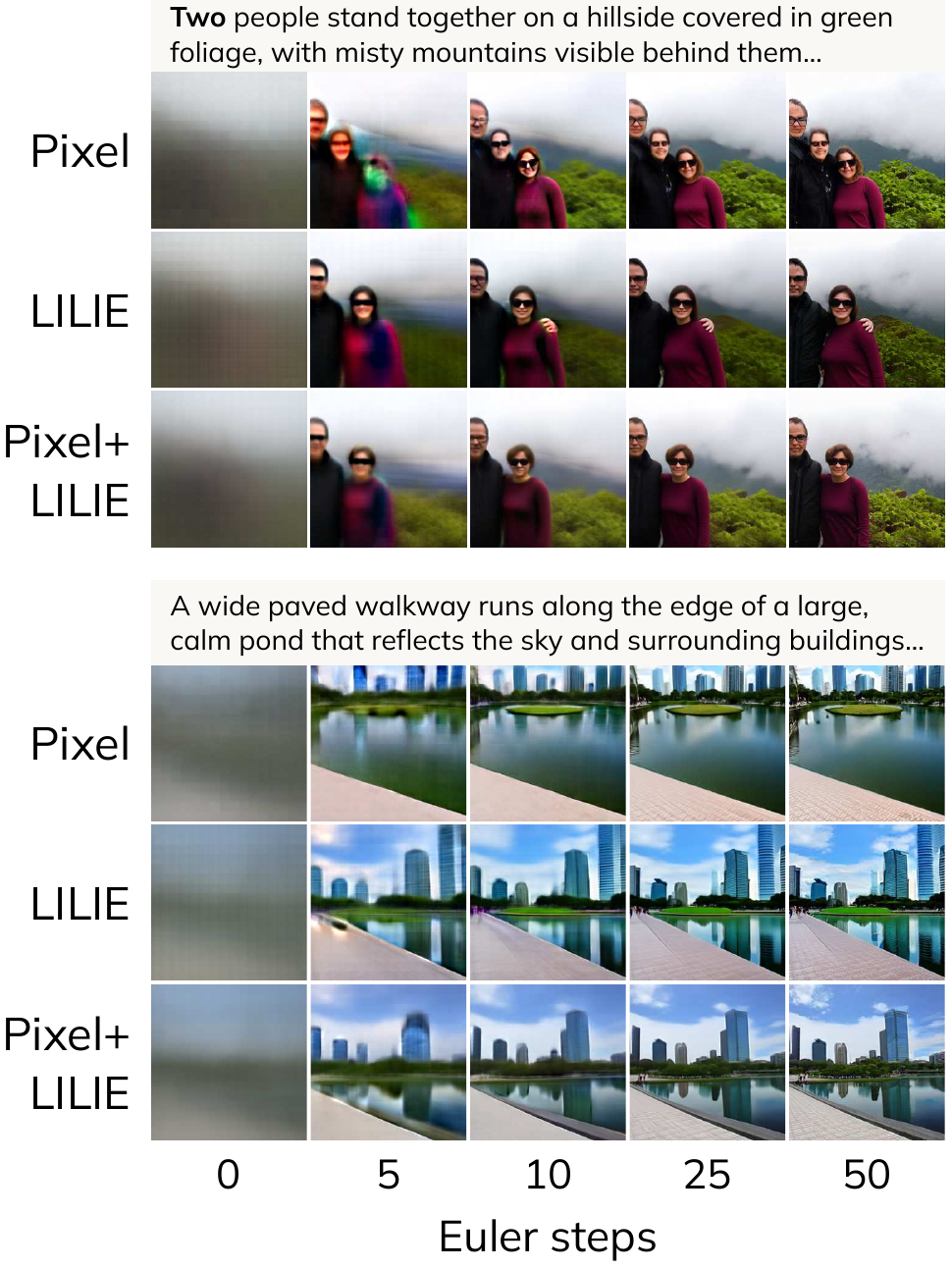}
\caption{\textbf{Selected sampling trajectories over 50 steps.} We fix a 50 step sampling budget and show clean image predictions at intermediate Euler steps 0, 5, 10, and 25, followed by the final image at step 50. Each example uses the same prompt and initial noise across methods.}
\label{fig:supp:sampling-trajectories}
\end{figure}

\begin{figure}[!ht]
    \centering
    \includegraphics[width=\linewidth,height=0.86\textheight,keepaspectratio]{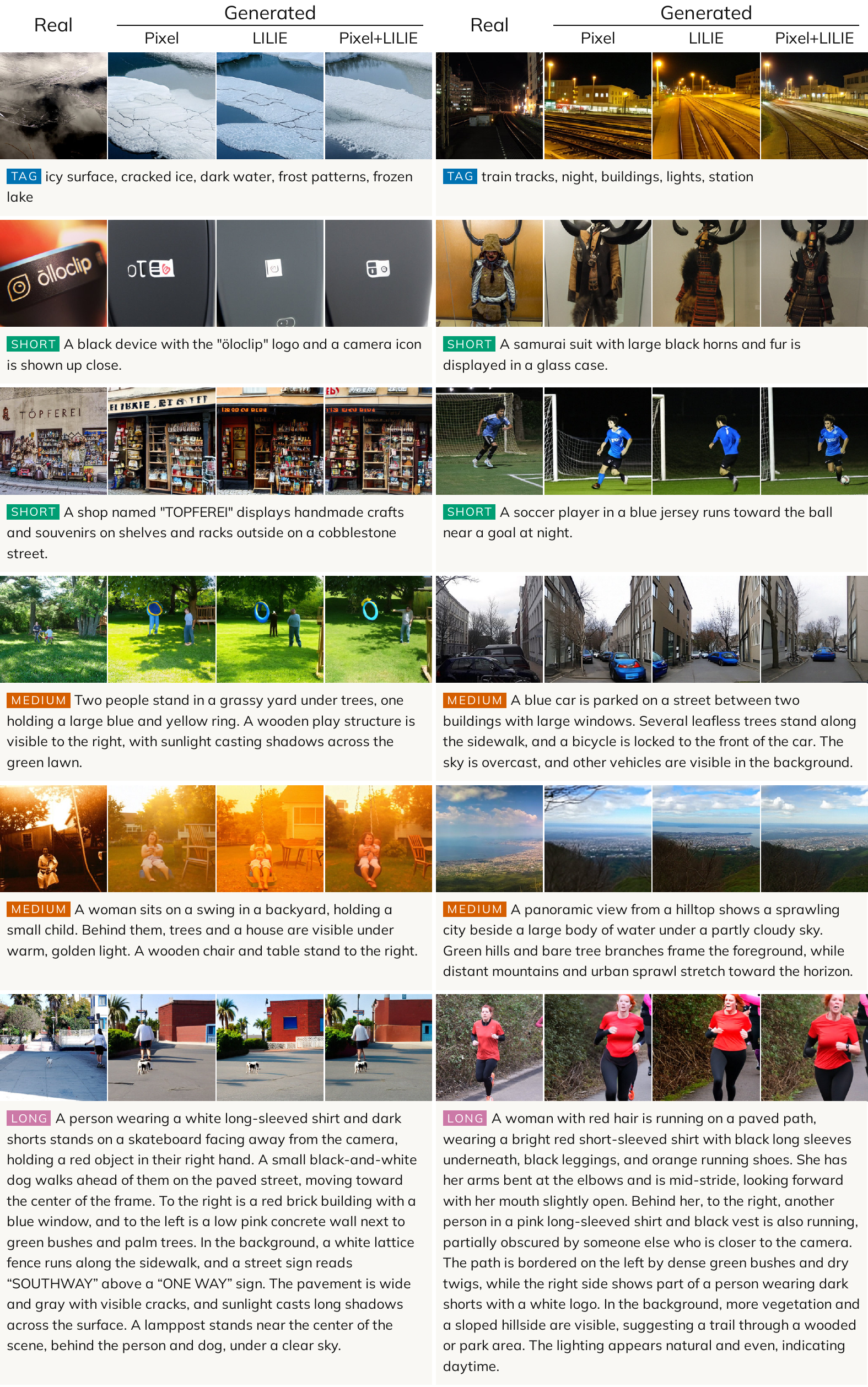}
    \caption{\uncuratedsamplescaption}
    \label{fig:supp:uncurated-samples-1}
\end{figure}

\begin{figure}[p]
    \centering
    \includegraphics[width=\linewidth,height=0.91\textheight,keepaspectratio]{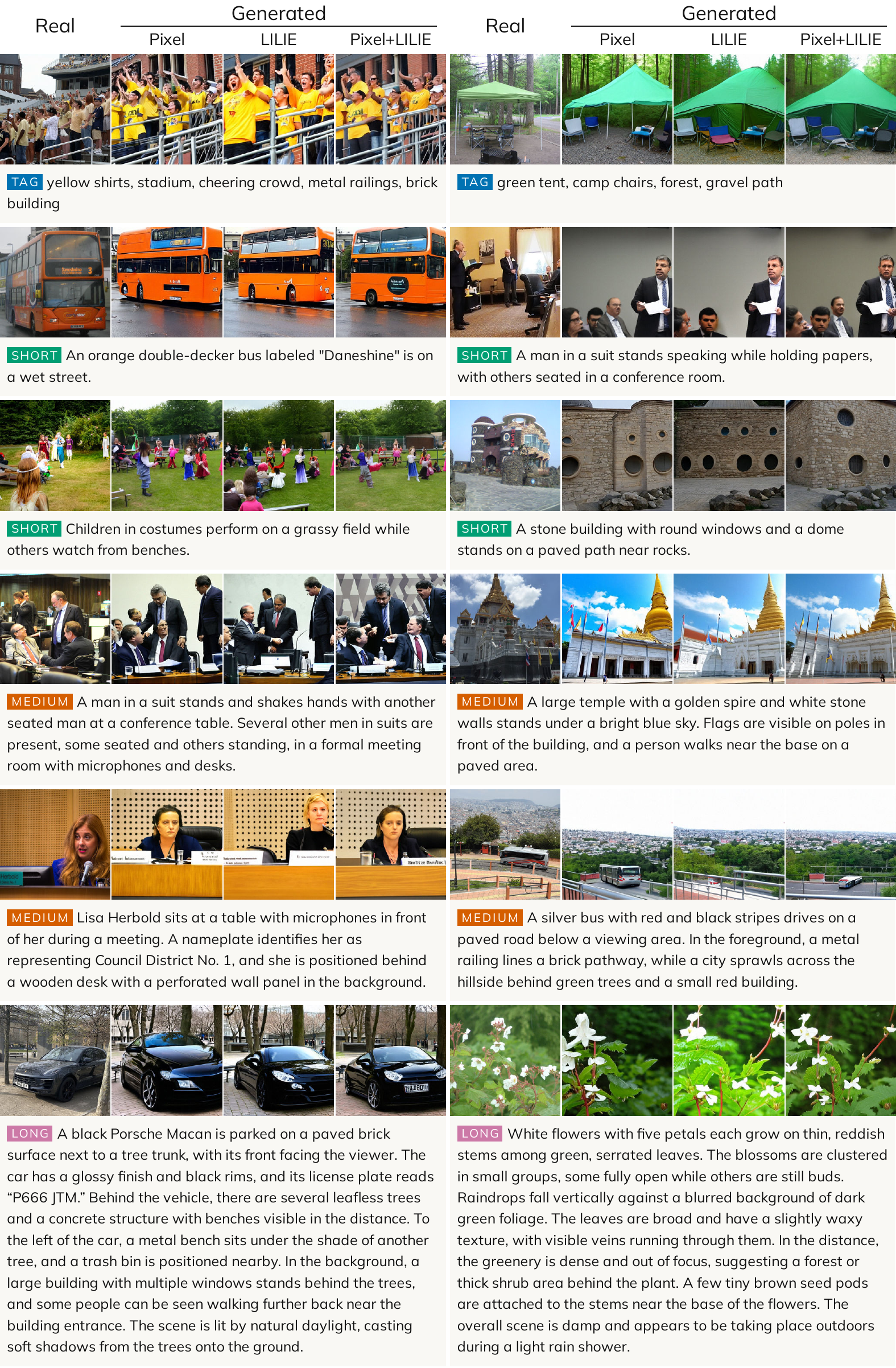}
    \caption{\uncuratedsamplescaption}
    \label{fig:supp:uncurated-samples-2}
\end{figure}

\begin{figure}[p]
    \centering
    \includegraphics[width=\linewidth,height=0.91\textheight,keepaspectratio]{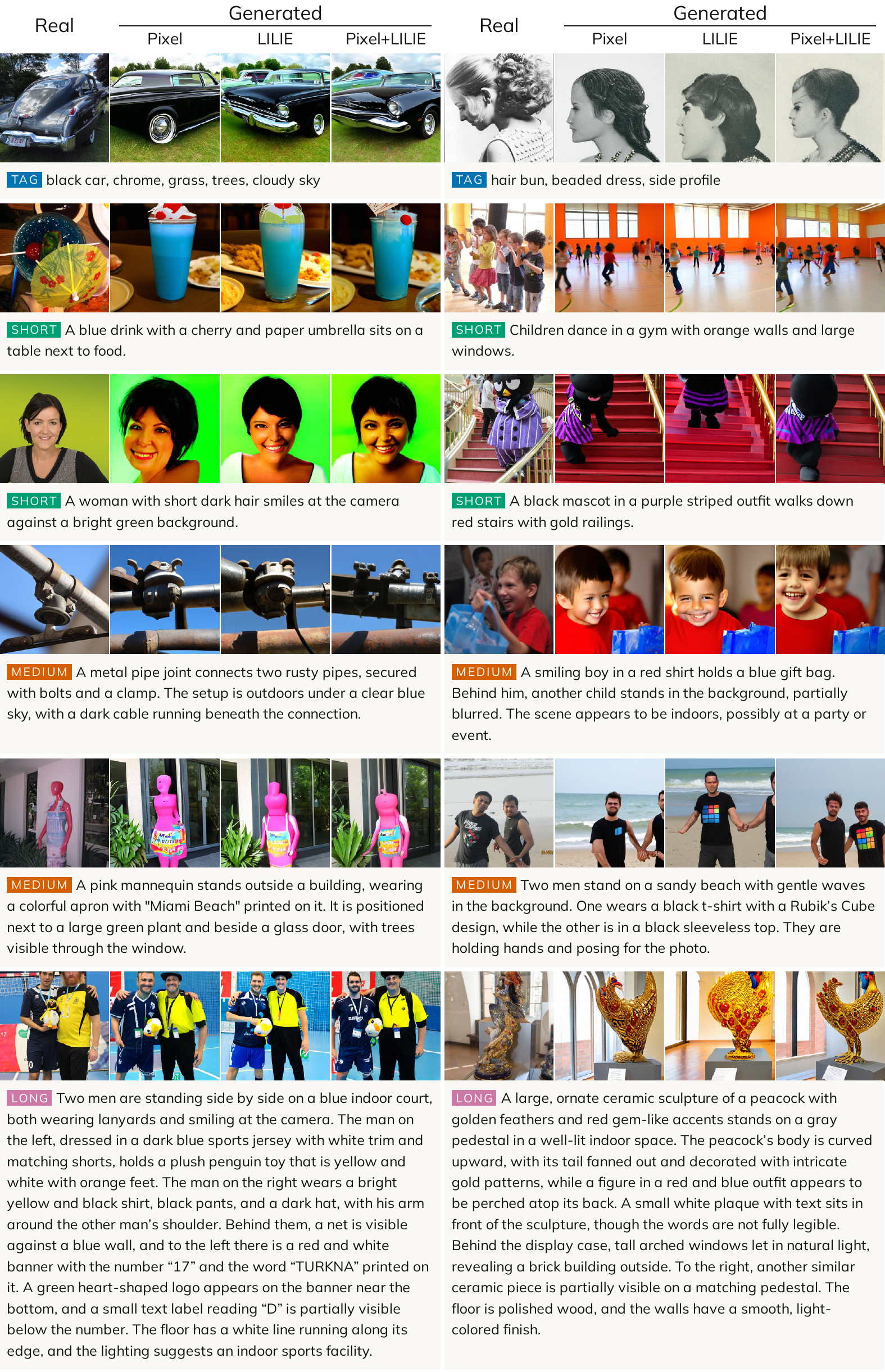}
    \caption{\uncuratedsamplescaption}
    \label{fig:supp:uncurated-samples-3}
\end{figure}

\begin{figure}[p]
    \centering
    \includegraphics[width=\linewidth,height=0.91\textheight,keepaspectratio]{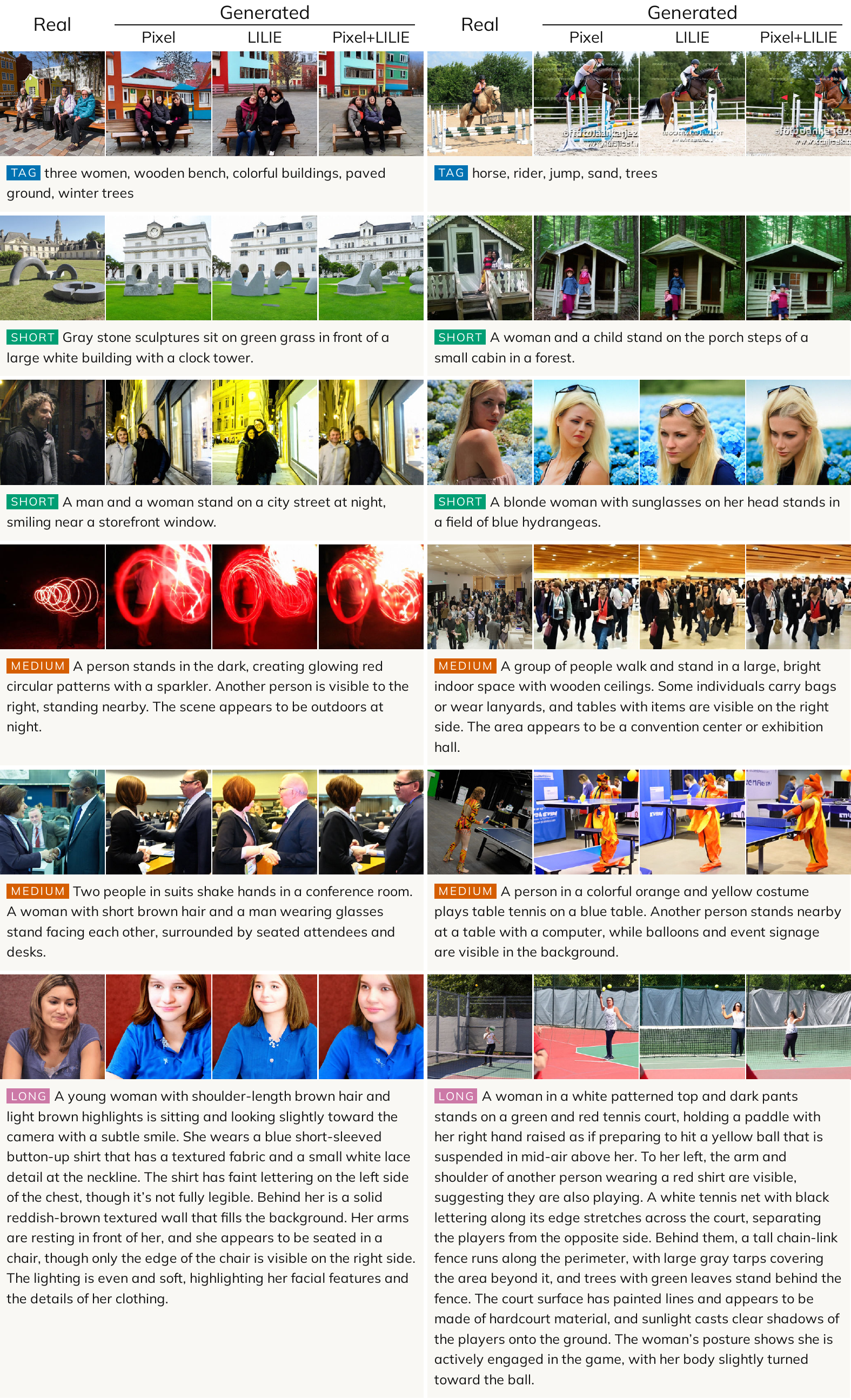}
    \caption{\uncuratedsamplescaption}
    \label{fig:supp:uncurated-samples-4}
\end{figure}

\begin{figure}[p]
    \centering
    \includegraphics[width=\linewidth,height=0.91\textheight,keepaspectratio]{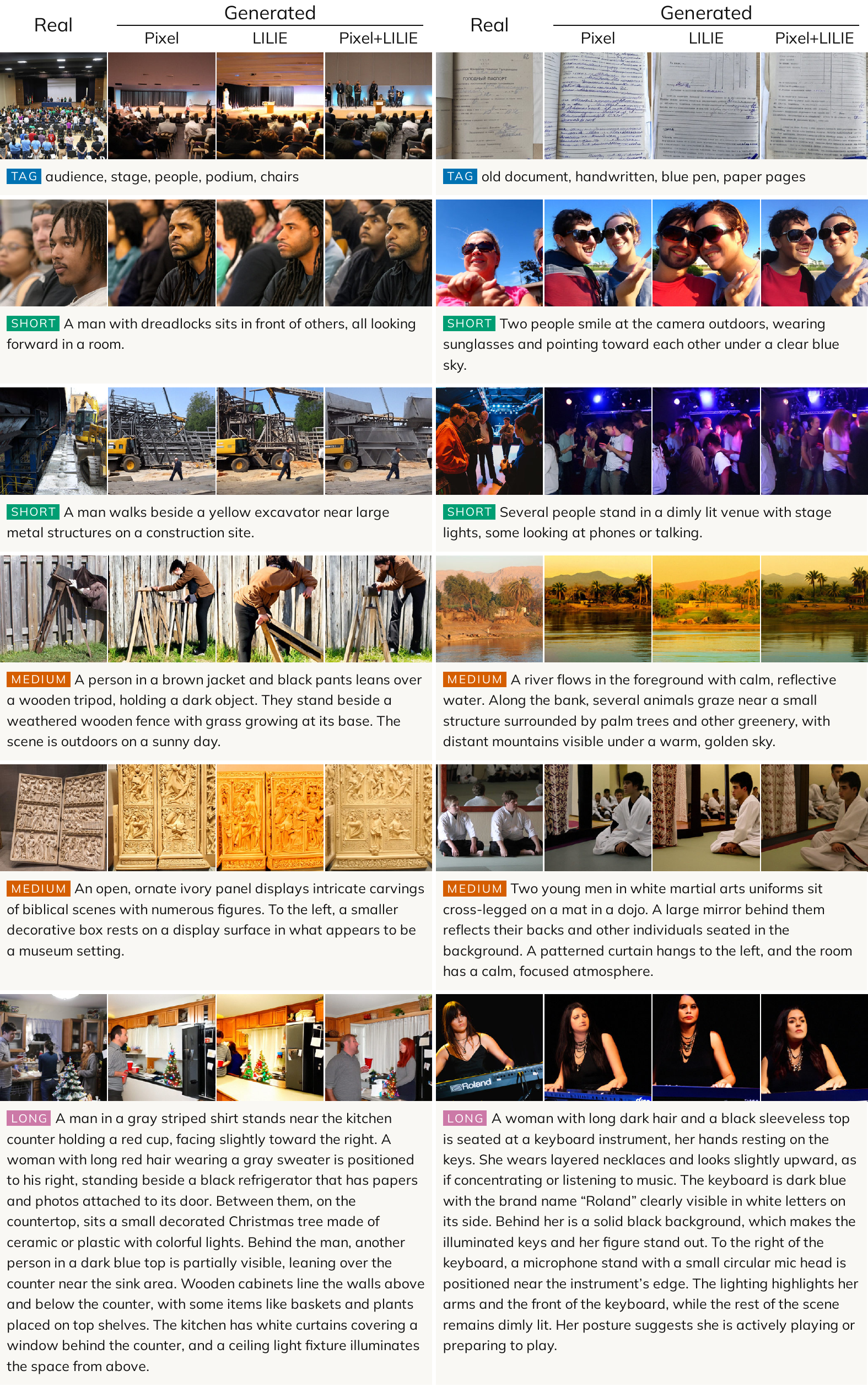}
    \caption{\uncuratedsamplescaption}
    \label{fig:supp:uncurated-samples-5}
\end{figure}

\begin{figure}[p]
    \centering
    \includegraphics[width=\linewidth,height=0.94\textheight,keepaspectratio]{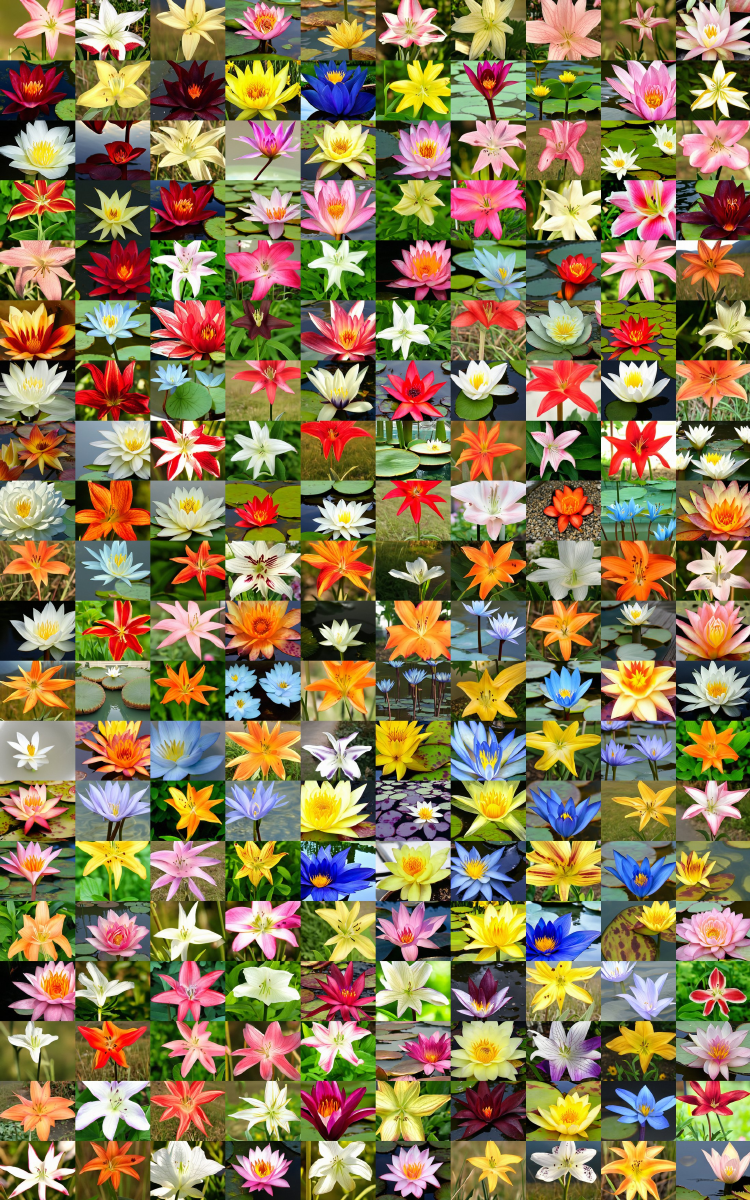}
    \caption{A mosaic of lilies and water lilies sampled from JiT-T2I-g trained with Pixel+\method{}.}
    \label{fig:supp:lily-flowers}
\end{figure}

We provide selected sampling trajectories for JiT-T2I-\gsize{} models trained with Pixel, \method{}, and Pixel+\method{} in Figure~\ref{fig:supp:sampling-trajectories}. We provide random samples from the same models in \Cref{fig:supp:uncurated-samples-1,fig:supp:uncurated-samples-2,fig:supp:uncurated-samples-3,fig:supp:uncurated-samples-4,fig:supp:uncurated-samples-5}. \Cref{fig:supp:lily-flowers} is a mosaic of selected samples of lily and water lily flowers from JiT-T2I-g trained with Pixel+\method{}. All samples are taken from each model's optimal CFG scale, listed in Table~\ref{tab:supp:headline-results-full}.

\clearpage

\clearpage

\clearpage

\clearpage

\end{document}